\documentclass{article} \usepackage{iclr2027_conference,times}

\usepackage{amsmath,amsfonts,bm}

\def\eqref#1{equation~\ref{#1}}

\def\1{\bm{1}}

\def\vx{{\bm{x}}}

\def\vz{{\bm{z}}}

\def\mX{{\bm{X}}}

\def\mZ{{\bm{Z}}}

\DeclareMathAlphabet{\mathsfit}{\encodingdefault}{\sfdefault}{m}{sl}
\SetMathAlphabet{\mathsfit}{bold}{\encodingdefault}{\sfdefault}{bx}{n}

\def\sT{{\mathbb{T}}}

\newcommand{\E}{\mathbb{E}}

\newcommand{\R}{\mathbb{R}}

\DeclareMathOperator*{\argmax}{arg\,max}

\usepackage[T1]{fontenc}
\usepackage[breaklinks]{hyperref}
\usepackage{url}
\usepackage{booktabs}
\usepackage{amsfonts} 
\usepackage{nicefrac}
\usepackage{xcolor}
\usepackage[table]{xcolor} 
\usepackage{color_edits}
\usepackage{float}
\usepackage{microtype}
\usepackage{upgreek}
\usepackage{amsbsy}
\usepackage{amssymb}
\usepackage{cancel}
\usepackage{amsmath}
\usepackage{amsthm}
\usepackage{graphicx}
\usepackage{setspace}
\usepackage{titlesec}
\usepackage{enumitem}
\usepackage{caption}
\usepackage{subcaption}
\usepackage{adjustbox,multirow}
\usepackage[noabbrev]{cleveref}
\usepackage{algorithm}
\usepackage{algpseudocode}
\usepackage{wrapfig}
\usepackage{titletoc}
\usepackage{makecell}

\hypersetup{
    colorlinks,
    linkcolor={red!100!black},
    citecolor={blue!100!black},
    urlcolor={blue!80!black}
}
\usepackage[most]{tcolorbox} 
\definecolor{orangesignal}{HTML}{FFAB0F} \definecolor{bluesignal}{HTML}{F2FCFF} \definecolor{redsignal}{HTML}{F77482} \definecolor{pinksignal}{HTML}{BE03FD} \newcommand{\firstbest}[1]{\textbf{\color{redsignal} #1}}
\newcommand{\secondbest}[1]{{\color{orangesignal} #1}}

\definecolor{citationcolor}{HTML}{017374} \definecolor{linkcolorcustom}{HTML}{017374}
\hypersetup{ 
    colorlinks=true,       
    linkcolor=pinksignal,
    citecolor=citationcolor, 
    urlcolor=linkcolorcustom 
}
\newcommand{\ours}{MoE-Learner}

\newtcolorbox{abox}{colback=bluesignal,
  colframe=citationcolor,
  boxrule=0.6pt,
  left=2pt, right=2pt, top=2pt, bottom=2pt
}

\newtheorem{theorem}{Theorem}

\title{Data-informed Adaptation of Foundation \\Models for Causal Effect Estimation}

\author{Chunwei Ma \& Russell D. Wolfinger \\
JMP Statistical Discovery \\
Cary, NC, USA \\
\texttt{\{Chunwei.Ma,Russ.Wolfinger\}@jmp.com}
}

\iclrfinalcopy
\begin{document}

\maketitle
\vspace{-5mm} 
\begin{abstract}
Due to the fundamental problem of causal inference -- the impossibility of observing counterfactual outcomes -- there is always a gap between the training goal (outcome, propensity) and the inference target (CATE, ATE). Traditional causal effect estimation revolves around setting assumptions, selecting identification strategies, and choosing causal estimators, while more recently, Causal Foundation Models (CFMs), built on Prior-Data-Fitted Networks (PFNs), ease this process by pretraining on massive synthetic prior datasets. Despite their success, such amortization makes causal inference less transparent, obscuring which assumptions the model implicitly relies on for a given dataset.
The high diversity of real-world data-generating processes (DGPs) and Structural Causal Models (SCMs) makes it hard to know, a priori, which causal inference strategy is compatible with a given dataset. In this work, we focus on this compatibility between an observational dataset and method-specific assumptions, and find that geometric characteristics of data embeddings -- specifically, the variogram -- provide a useful signal of whether a given causal estimation strategy is appropriate. We develop MoE-Learner, a mixture-of-causal-experts framework that embeds a dataset under multiple assumption-specific views and uses the variogram as a counterfactual-free router to assess each expert's validity. Rather than eliminating causal assumptions, this shifts their validation from a fixed, a priori choice to a criterion that can be checked directly from data geometry. MoE-Learner achieves the best average rank on benchmark datasets (IHDP, ACIC, Lalonde).
Being orthogonal to (causal) foundation models, our method builds a bridge from data characteristics to causal estimation strategy rather than proposing another CFM. \end{abstract}
\vspace{-5mm} 
\section{Introduction} \label{sec:introduction}
\begin{wrapfigure}{r}{0.38\textwidth} 
    \vspace{-2mm}
    \centering
    \subfloat{\includegraphics[width=0.38\textwidth]{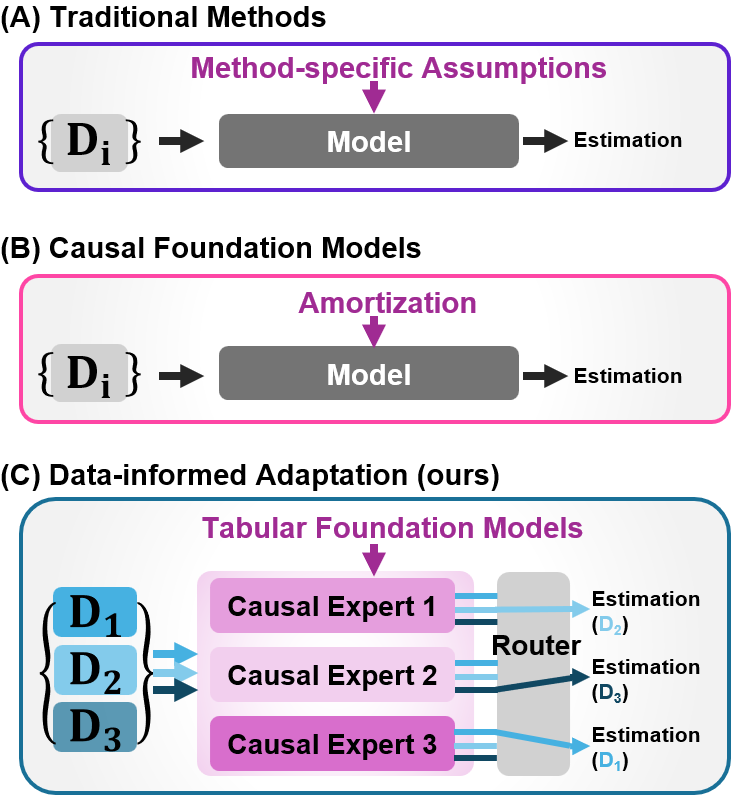}} 
\caption{Comparison of different paradigms.}
	\label{fig:compare}
    \vspace{-5mm} 
\end{wrapfigure}
The scientific exploration of the world can be summarized in three steps: (1) finding connections between objects (correlation); (2) identifying causation within these correlations (causal inference)~\citep{pearl2009causality,pearl2018book}; and (3) mechanistic explanation~\citep{machamer2000thinking,craver2007explaining,woodward2005making}. Artificial Intelligence (AI), especially Deep Learning (DL) and transformer-based Large Language Models (LLMs), has achieved remarkable success in correlation prediction -- e.g., linking images to their labels or predicting the next word -- but a gap between correlation and causation remains~\citep{kiciman2023causal,scholkopf2021toward}.

Causal effect estimation aims at identifying whether an action (e.g., an educational training or a drug treatment) has a causal effect on a population (e.g., income after training, or survival rate after treatment)~\citep{rubin1974estimating, rosenbaum1983central}. Traditionally, assumptions play a central role in causal inference, where bespoke estimators are tailored to specific settings in which certain assumptions hold. Beyond the common assumptions of Unconfoundedness/Exchangeability, Positivity/Overlap, and SUTVA~\citep{rubin1980randomization, rosenbaum1983central}, each estimator typically carries additional, method-specific assumptions: for example, S-Learner assumes the response surfaces for treatment ($Y_1$) and control ($Y_0$) share a common functional form, whereas T-Learner models them as distinct functions~\citep{kunzel2019metalearners}. Such an assumption need not hold for a given dataset, and a mismatch can lead to a biased estimate~\citep{athey2016recursive}. 

Traditional causal inference therefore consists of two steps~\citep{pearl2009causality, rubin2008objective}:
(I) Causal modeling: identifying cause-and-effect mechanisms, isolating confounding, and making assumptions~\citep{pearl2010introduction, rosenbaum1983central}.
(II) Model specification: choosing estimators, determining functional forms, and tuning hyperparameters~\citep{kunzel2019metalearners, chernozhukov2018double}.
\begin{figure}
    \centering
    \subfloat{\includegraphics[width=\textwidth]{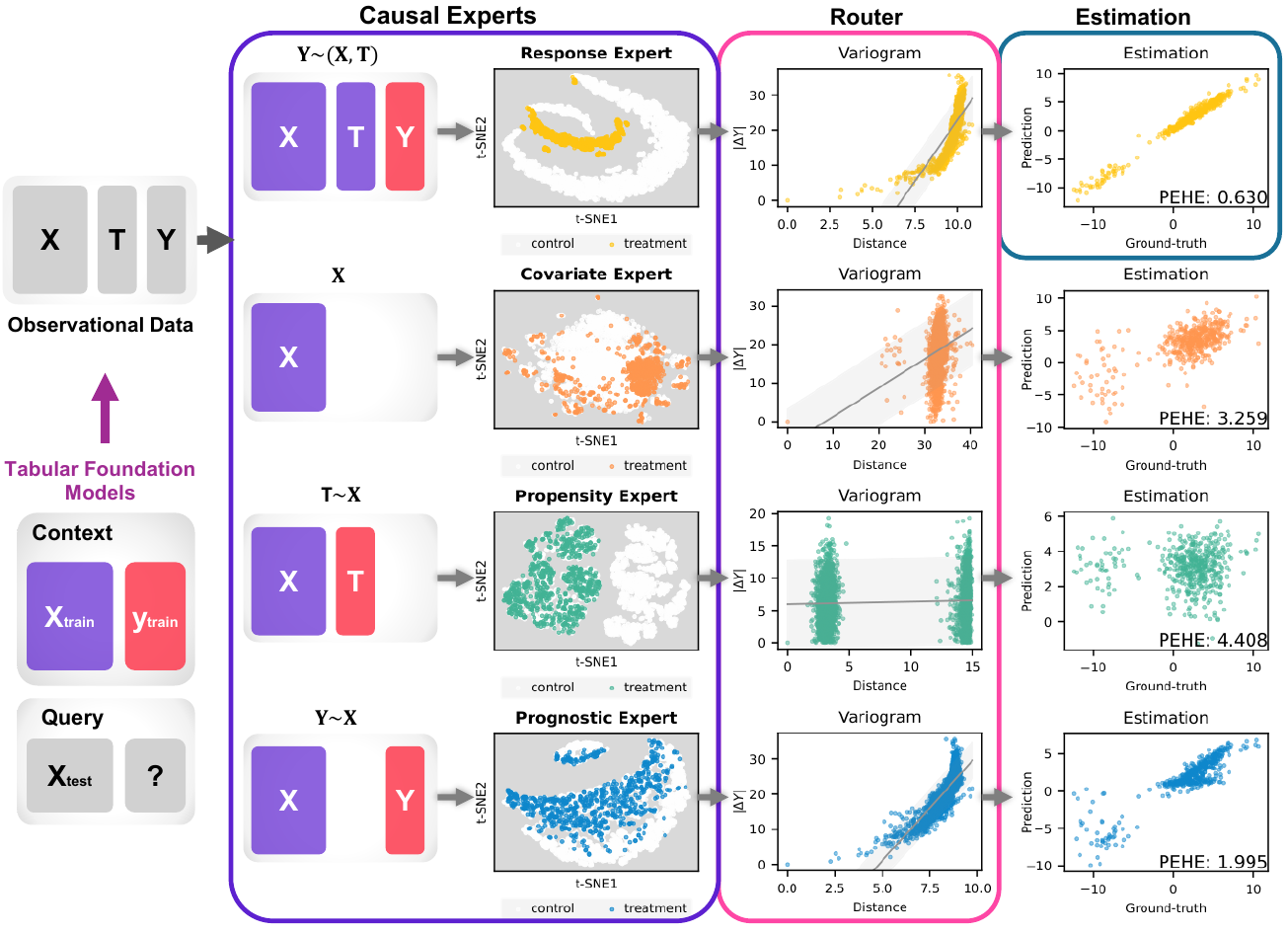}} 
    \caption{Illustration of the \ours{} workflow using ACIC data as the example. The observational data $(\mX, T, Y)$ is converted to four kinds of embeddings by modeling different relationships ($Y \sim (\mX, T)$, $Y \sim \mX$, $T \sim \mX$, and a covariate modeling of $\mX$). Each expert makes its own causal estimation. The Router selects the expert by evaluating the variograms of the embeddings.}\label{fig:workflow}
    \vspace{-2mm}
\end{figure}
Recently, Prior-Data-Fitted Networks (PFNs)~\citep{hollmann2023tabpfn,helli2024drift}, e.g., TabPFN~\citep{hollmann2025accurate} and TabICL~\citep{qu2025tabicl}, have introduced a groundbreaking new paradigm for tabular learning, in which a transformer is pretrained on massive synthetic tables, achieving consistently strong performance across diverse tabular data and tasks~\citep{nagler2023statistical,adriaensen2023efficient}. Inspired by PFNs, Causal Foundation Models (CFMs)~\citep{stith2026causal} have since been proposed -- e.g., CausalPFN~\citep{balazadeh2026causalpfn}, CausalFM~\citep{ma2025foundation}, and Do-PFN~\citep{robertson2025pfn} -- which are pretrained on diverse data-generating processes (DGPs) and Structural Causal Models (SCMs), and then perform inference for a new dataset by integrating over the resulting Bayesian posterior.

However, evidence suggests that CFMs, which estimate the treatment effect directly from an input dataset in a single forward pass, bypassing step-by-step intermediate mechanisms such as propensity weighting (IPW) or g-computation, can perform suboptimally -- particularly on out-of-distribution real-world medical data that differs substantially from the prior distribution of SCMs used in training~\citep{mourao2026priordatafittednetworkscausal}.

In this work, we ask the following research questions regarding current challenges for causal effect estimation:
(RQ1) Each estimator requires a set of model-specific assumptions. However, do they apply to the data? To what extent?
(RQ2) When settling on causal assumptions, which estimation strategy should be selected? Can a strategy adapt flexibly to the data?
(RQ3) In practice, how can a general-purpose tabular foundation model be retrofitted into a causal estimator with minimal causal assumptions?

In the following sections, we first reveal the connections from causal inference to Prior-Data-Fitted Networks, Mixture of Experts, and Test-Time Adaptation, and then introduce a principled framework.

\section{Background and Related Work} \label{sec:related}
\textbf{Causal Effect Estimation.}
Many causal effect estimation methods have been developed over the past several decades. Beyond common assumptions such as unconfoundedness and positivity, each method typically relies on additional assumptions, which we refer to as method-specific assumptions. For example, IPW~\citep{rosenbaum1983central} relies entirely on a correctly specified propensity model and is highly sensitive to positivity violations. In contrast, methods such as RA-Net~\citep{curth2021inductive} and BART~\citep{hill2011bayesian} perform direct regression adjustment without any propensity correction, leaving them exposed to covariate shift between treatment arms. Other methods, such as DragonNet~\citep{shi2019adapting} and TarNet~\citep{shalit2017estimating}, jointly model both arms through a shared representation and incorporate propensity estimation, implicitly assuming that a representation sufficient for predicting treatment is also sufficient for predicting the outcome.
While these varied and often sophisticated assumptions pose significant challenges in selecting a proper method for a real-world dataset, formulating a coherent and correct set of assumptions itself requires profound domain expertise~\citep{pearl2009causality, imbens2015causal}. More importantly, once a set of assumptions is fixed, whether it aligns with the observed data -- and to what extent -- is usually left unexamined~\citep{kang2007demystifying, parikh2022validating}, so the strategy remains fixed without adaptation to the real-world data.
\begin{abox}
\textbf{Key Takeaway:} For traditional causal inference methods, method-specific assumptions are typically tailored to the method rather than adaptive to the data.
\end{abox}

\textbf{Algorithm Selection for Causal Inference.}
A separate line of work addresses this fixation by treating estimator choice itself as a selection problem~\citep{schuler2018comparison, nie2021quasi, parikh2022validating}. Since CATE estimators cannot be validated against ground-truth counterfactuals, prior work instead relies on \emph{surrogate metrics}, such as R-scorer or doubly-robust proxy losses, computed on a held-out validation set of observed outcomes~\citep{foster2023orthogonal, lan2024causal}: ~\citet{mahajan2024empirical} benchmark 34 such metrics, using them to select among CATE estimators based on their agreement with held-out observed outcomes. This approach, however, still depends on a labeled validation split, and in small or imbalanced causal datasets, setting one aside can materially reduce the effective sample size available for estimating treatment effects, particularly for the smaller treatment arm. 

\textbf{Foundation Models for Causal Effect Estimation.}
Different causal foundation models amortize estimation over a prior of SCMs with varying assumptions to ease causal inference at test time~\citep{stith2026causal,bellot2026foundationmodelspartialcausal}. CausalPFN~\citep{balazadeh2026causalpfn} assumes the user has specified an identification regime restricted to the back-door criterion; CausalFM~\citep{ma2025foundation} generalizes this to back-door, IV, and front-door regimes. In contrast, Do-PFN~\citep{robertson2025pfn}, MACE-TNP~\citep{dhir2026estimating}, and ACTIVA~\citep{sauter2025activa} instead specify a broad prior spanning many plausible SCMs. However, this amortization makes causal inference less transparent: among the SCMs in the pretraining prior, which align well with a given test dataset, which diverge from it, and how should such disagreements be reconciled? A prior that is too broad risks poor specificity, while one that is too narrow (e.g., models trained specifically for front-door or IV settings) risks poor coverage of real-world regimes. \citet{reuter2026use} address this by conditioning on prior causal information such as partial ancestral graphs, though such information is not always available.

\textbf{Mixture of Experts for Causal Inference.}
Given heterogeneous causal structures, mixed assumptions, and diverse estimator choices, the mixture-of-experts (MoE) architecture is a natural fit~\citep{shazeer2017outrageously, kunzel2019metalearners}. In LLMs, MoE learns a dispatching mechanism (the router) that assigns each token to an "expert" -- a dense subnetwork -- enabling both test-time acceleration and smart allocation to the expert best suited to the token (e.g., grammar, math, coding)~\citep{fedus2022switch}. Adapting MoE to causal effect estimation, however, is not straightforward: unlike next-token prediction, which provides a self-supervised training signal for the router, the fundamental problem of causal inference -- the impossibility of observing both treated and untreated outcomes for the same unit -- makes it impossible to train the router and the experts jointly, end-to-end, for CATE/ATE prediction. CausalMoE~\citep{liu2026causalmoe}, a foundation model for multimodal Granger causal discovery, designs experts for heterogeneous modalities in time-series causal discovery (e.g., semantic, visual, temporal, and frequency experts), but it remains unclear how to adapt MoE to the differing, and possibly unknown, causal structures and regimes relevant to effect estimation.

\textbf{Test-Time Adaptation of Pretrained Models.}
Deep neural networks often suffer significant performance degradation under domain shift. Test-Time Adaptation (TTA) methods address this by adapting a trained model to the test domain without labeled test data or access to information from the training procedure (e.g., the training dataset or its feature statistics)~\citep{sun2020test,wang2020tent,schneider2020improving,gozeten2025test}.
Since PFNs are pretrained solely on synthetic simulation data, a distribution shift between the prior's DAGs or SCMs and real-world ones is expected. We identify causal effect estimation as fundamentally akin to TTA: an ideal model, or meta-model, must select a proper identification strategy or estimator without knowing the ground-truth CATE/ATE, the true underlying DGP or SCM, or having access to the pretraining prior. In TTA, since no ground-truth label is available at test time, methods instead exploit data-dependent information such as nearest-neighbor structure~\citep{jang2022test} and local geometry~\citep{lei2025ttvd}. Recent work has moved toward prior-free tabular TTA~\citep{he2026prior,lee2026mitigating}; however, the fundamental problem of causal inference renders such methods inapplicable here, since they address a shift between training and test distributions \emph{for the same predictive task}, not the more basic discrepancy between the training goal (outcome or propensity prediction) and the testing goal (CATE/ATE estimation).

\vspace{-2mm}
\section{Overview of Contributions} \label{sec:contribution}
In real-world datasets, the DGPs and SCMs are highly diverse, making it unclear which causal inference strategy to use. We address this in two steps, corresponding to our research questions. First (RQ3), we unify and categorize causal inference strategies into four modeling types: response modeling ($Y \sim (X, T)$), propensity modeling ($T \sim X$), prognostic modeling ($Y \sim X$), and a covariate embedding ($X$). We design a "causal expert" for each modeling type: for the first three, an expert places the corresponding variables into the context of a tabular foundation model and extracts an embedding vector for each unit; the covariate expert instead uses a standardized embedding of $X$ itself -- a label-free, RAG-like reference view, rather than one learned through the tabular foundation model. Each expert therefore not only converts units into embeddings, but also implicitly encodes the assumptions underlying its modeling choice (e.g., the propensity expert assumes that treatment assignment is related to the covariates $X$). This retrofits a general-purpose tabular foundation model into a suite of causal estimators, each carrying minimal, explicit assumptions.

While the causal experts encapsulate multiple causal inference strategies, the next challenge is how to route to the proper expert for a given dataset (RQ1, RQ2). Unlike traditional MoE, we lack a direct training signal for the router: there is no observation of counterfactual outcomes, no true CATE/ATE performance for any expert, and no separate validation set. To address this, we borrow the idea of the variogram from spatial statistics -- a function measuring how much sample labels differ from one another as their relative distance increases. Analogously, if a causal expert's embedding is valid, nearby units in that embedding space should also share similar outcome values; we use this local smoothness as a ground-truth-free diagnostic of how well an expert's implicit assumptions fit the data (RQ1), and route to the expert whose embedding best satisfies it (RQ2). Empirically, we find that different experts are activated for different datasets.

The framework relies on standard causal identification assumptions, but shifts the choice among method-specific estimation strategies from a fixed, a priori commitment to a criterion determined by the geometry of the joint data-assumption embeddings. We do not rely on empirical performance on a held-out validation set; instead, our routing criterion is grounded in a statistical property of the embedding space itself, making the selection more directly interpretable than validation-based alternatives. 

Our method is orthogonal to causal foundation models: we do not develop yet another CFM (Figure~\ref{fig:compare}). Instead, we show how to adapt a general-purpose tabular foundation model into a flexible causal effect estimator via causal experts and a router without counterfactual supervision, achieving better performance and improved interpretability.
 \def\featureDim{n}

\section{Methodology} \label{sec:method}
\subsection{Preliminary: Causal Effect Estimation} \label{sec:preliminary}
We work within the potential outcomes framework~\citep{rubin2005causal} and observe samples of random variables $(X, T, Y)$, where $X \in \R^\featureDim$ are the observed covariates with dimension $\featureDim$, $T$ is the treatment assignment drawn from a finite treatment set (binary treatments $T \in \sT = \{0, 1\}$ here), and $Y \in \R$ is the outcome of interest. For every $t \in \sT$, $Y_t \in \R$ is the potential outcome under treatment $t$, while the observed (factual) outcome is $Y$. The observational data $\{x, t, y\}$ are drawn from an unknown joint distribution $P(X, T, \{Y_t\}_{t \in \sT})$.

Causal inference typically requires the propensity (treatment assignment) distribution $\pi_t(x)$ and the expected potential outcomes $\mu_t(x)$, commonly referred to together as the nuisance parameters $\eta = (\mu_0, \mu_1, \pi)$:
\begin{align}
\pi_t(x) &= P[T = t \mid X=x], \\
\mu_{t}(x) &= \E [Y_t \mid X=x].
\end{align}

Our targets of inference are two quantities: the conditional average treatment effect (CATE), which represents the average effect of intervention ($Y_1 - Y_0$) on the subpopulation with covariates $X = x$, and the average treatment effect (ATE), which measures the average difference in outcomes between the entire population receiving treatment versus the same population not receiving it:
\begin{align}
    \text{CATE}: \quad & \tau(x) = \E[ Y_1 - Y_0 \mid X = x ]  = \mu_1(x) - \mu_0(x), \label{eq:cate} \\
    \text{ATE}: \quad & \lambda  = \E[ Y_1 - Y_0 ]  = \E[ \mu_1(X) - \mu_0(X) ]. \label{eq:ate}
\end{align}

\subsection{Causal Experts} \label{sec:experts}
\textbf{Causal Embeddings via Tabular Foundation Models.}
Unlike an embedding from a standard deep neural network, where the embedding vector is simply the penultimate-layer activation, extracting an embedding $Z = g(Q; C)$ from a Prior-Data-Fitted Network -- the contextualized hidden state for query $Q$ after the transformer layers $g$ -- requires specifying a context $C$~\citep{ye2025closer}. For example, to extract the embedding $\mZ_{XTY}$ from a TabPFN model predicting $Y$ from the concatenation of the dataset matrix $\mX$ and $T$, the query is $Q = [\mX, T]$ and the context is $C = ([\mX, T], Y)$, giving $\mZ_{XTY} = g([\mX, T]; ([\mX, T], Y))$.
Different choices of query and context correspond to different modeling relationships underlying traditional causal estimators. We exploit this correspondence to design four causal experts, each built from a different query-context pairing:\\
(1) \textbf{Response Expert} ($\mZ_{XTY} = g([\mX, T]; ([\mX, T], Y))$): As in S-Learner, BART, DragonNet, RA-Net, and TarNet, treatment is modeled as a covariate within an outcome model, i.e., $Y \sim F(\mX, T)$. The response expert embeds this relationship by taking $[\mX, T]$ as the query and $Y$ as the context, yielding a contextualized representation that jointly encodes covariates, treatment assignment, and their interaction with the outcome.\\
(2) \textbf{Propensity Expert} ($\mZ_{XT} = g(\mX; (\mX, T))$): As in IPW, and as a component within DA-Learner, X-Learner, DragonNet, GRF, and Forest DML, a propensity model is estimated to reweight or adjust samples according to $T \sim F(\mX)$. The propensity expert embeds this relationship by taking $\mX$ as the query and $T$ as the context, producing a representation of each unit's position within the treatment-assignment mechanism, independent of the outcome.\\
(3) \textbf{Prognostic Expert} ($\mZ_{XY} = g(\mX; (\mX, Y))$): As in T-Learner, X-Learner's first stage, DA-Learner, and the prognostic function in BCF, an outcome model $Y \sim F(\mX)$ is estimated to capture the relationship between covariates and outcome, pooling all samples into a shared representation space.\\
(4) \textbf{Covariate Expert} ($\mZ_X = \frac{\mX - \mu_X}{\sigma_X}$): The standardization of the input covariates $\mX$, containing no information about $T$ or $Y$. This plays a role analogous to the embedding step in retrieval-augmented generation (RAG): just as RAG embeds a query to retrieve semantically similar documents, the covariate expert embeds each unit to retrieve similar units in covariate space, providing a matching-style signal that complements the outcome- and treatment-aware experts above.

For each query unit $\vx$, since we want to predict potential outcomes under both control and treatment, each unit is embedded into two representations, $\vz_0$ and $\vz_1$. For the Response Expert, these two embeddings differ: $\vz_{t} = g([\vx, t]; ([\mX, T], Y)), t \in \{0,1\}$. For the other three experts, $\vz_0$ and $\vz_1$ share the same embedding, since their query does not depend on $t$.

\begin{abox}
\textbf{Key Takeaway:} We embed a dataset under several causal experts in parallel, so that no single method-specific assumption is committed to a priori, while each expert individually remains minimally assumption-laden.
\end{abox}

\textbf{Top-K Attention from Query to Context.} \label{sec:topk}
We refer to the generated embeddings for the TabPFN query $Q$ and context $C$ as the query and context embeddings, respectively. For each expert, given a context set $\mZ_t = \{\vz_t^{(i)}\}_{i=1}^{N_t}$ of feature-label pairs, where $t \in \{0,1\}$ denotes the control/treatment group, we cast CATE prediction as an attention-based retrieval process. For a query embedding $\vz_t$, we define a feature dissimilarity $D$ between points $\vz_t$ and $\vz_t^{(k)}$ (in practice, cosine distance), and retrieve the top-$K$ most similar context pairs $\{(\vz_t^{(k)}, y_t^{(k)})\}_{k=1}^{K}$ -- i.e., the $K$ context points with smallest dissimilarity $D(\vz_t, \vz_t^{(k)})$ to the query. Attention scores over the retrieved neighborhood are assigned by a learnable weighting function $f_\theta: \mathbb{R} \to \mathbb{R}_{>0}$, implemented as a small MLP with Softplus activations and non-negative weight matrices, and normalized over the retrieved set:
\begin{footnotesize}
$$
a_t^{(k)} = \frac{f_\theta\big(D(\vz_t, \vz_t^{(k)})\big)}{\sum_{j=1}^{K} f_\theta\big(D(\vz_t, \vz_t^{(j)})\big)}, \qquad k = 1, \dots, K.
$$
\end{footnotesize}
The query label is then predicted as the attention-weighted average of the retrieved labels,
\begin{footnotesize}
$$
\hat{y}_t = \sum_{k=1}^{K} a_t^{(k)} \, y_t^{(k)}.
$$
\end{footnotesize}
Unlike standard self-attention, we omit learned query/key projections: scores are computed directly from a fixed dissimilarity between raw feature embeddings, with $f_\theta$ as the only learned component of the attention mechanism.
Finally, the CATE for unit $\vx$ is predicted as
$$
\hat\tau(\vx) = \hat{y}_1 - \hat{y}_0.
$$

\subsection{Causal Router} \label{sec:router}
\textbf{Variogram: Data-Informed Evaluation of Expert Validity.}
When different experts implicitly encode different causal relationships, the natural next question is which expert to trust for a given dataset. Rather than committing a priori to a single, universally preferred causal estimator, \ours{} assesses each expert's validity directly from the geometry of its own embedding space. We borrow the idea of the \emph{variogram} from geostatistics: a function that measures how the expected squared difference between two labels grows as the distance between their locations increases. Applied to a causal expert's embedding, the analogous question is whether units that are close in that expert's embedding space also share similar outcomes. If an expert's underlying assumption is well matched to the data, embedding proximity should track outcome similarity, so nearby units exhibit small label variation, and this variation should not shrink as embedding distance grows. An expert whose embedding shows no such relationship -- where outcome similarity is unrelated to embedding proximity -- encodes an assumption that fails to organize the data in a causally meaningful way, and should be trusted less.

\textbf{Theoretical Analysis.}
We formalize this intuition for a single expert $e$ and treatment group $t$; the group subscript $t$ is suppressed below, as the analysis applies identically and independently to each group's embeddings. Let $\vz^{(e)}$ denote the embedding produced by expert $e$, and let $f_e(\vz) = \E[Y \mid \vz^{(e)} = \vz]$ be the regression function it induces. We model
$$
Y = f_e(\vz^{(e)}) + \varepsilon, \qquad \E[\varepsilon \mid \vz^{(e)}] = 0, \quad \operatorname{Var}(\varepsilon) = \sigma^2,
$$
and let $D$ be the dissimilarity used for neighbor retrieval in $\vz^{(e)}$-space (Section~\ref{sec:topk}). We say $f_e$ is \emph{$\beta$-locally smooth} with respect to $D$ if there exists $L_e > 0$ such that $|f_e(\vz) - f_e(\vz')| \le L_e D(\vz, \vz')^\beta$ for all $\vz, \vz'$ in the relevant support. For an independent pair $(\vz_i^{(e)}, Y_i), (\vz_j^{(e)}, Y_j)$ drawn from the same distribution, define the {distance-label variogram}
$$
\gamma_e(d) = \tfrac12\, \E\big[(Y_i - Y_j)^2 \mid D(\vz_i^{(e)}, \vz_j^{(e)}) = d\big].
$$

\begin{theorem}[Distance-Label Variogram Bound]
\label{thm:variogram}
Under the model above, if $f_e$ is $\beta$-locally smooth with respect to $D$ with constant $L_e$, then for all $d$ in the support of $D$,
$$
\sigma^2 \;\le\; \gamma_e(d) \;\le\; \sigma^2 + \tfrac12 L_e^2 d^{2\beta}.
$$
\end{theorem}
The proof (Appendix~\ref{app:proof}) decomposes $\gamma_e(d)$ into an irreducible noise term $\sigma^2$ and a term governed by the local variation of $f_e$, then bounds the latter using the smoothness condition. The theorem implies that as $d \to 0$, $\gamma_e(d)$ is squeezed toward the noise floor $\sigma^2$: for a valid expert, near-duplicate units in embedding space must have near-identical expected outcomes, up to noise. Note that the bound constrains the \emph{ceiling} on label variation, not the exact shape of $\gamma_e$; it does not by itself force $\gamma_e(d)$ to increase in $d$. We therefore treat an increasing empirical trend between $d$ and label variation as evidence \emph{consistent with} local smoothness -- and hence with a valid expert -- rather than as a quantity derived directly from the theorem.

\textbf{Correlation as an Empirical Surrogate for Expert Validity.} \label{sec:surrogate}
Theorem~\ref{thm:variogram} characterizes a population relationship between distance and label variation; it does not by itself yield a single scalar for comparing experts. As an empirical surrogate, for each expert $e$ and group $t$ we sample $M$ anchor points from the context set $\mZ_t^{(e)}$ via farthest-point sampling. For each anchor, we compute its embedding distance $D$ and absolute label difference $|\Delta y|$ to every other point $i$ in $\mZ_t^{(e)}$, summarize their relationship with the Spearman rank correlation, and average these pairs across all $M$ anchors
$$
\rho_e^{(t)} = \tfrac{1}{M} \textstyle \sum_{k=1}^M \mathrm{corr}\big(\{d_{ki}\},\, \{|\Delta y_{ki}|\}\big).
$$
A large, positive $\rho_e^{(t)}$ indicates that label variation tracks embedding distance -- consistent with the local smoothness Theorem~\ref{thm:variogram} predicts for a valid expert -- while $\rho_e^{(t)} \approx 0$ indicates that expert $e$'s embedding carries little information about outcome similarity for group $t$. The router selects, independently per group,
$$
e_t^\star = \argmax_{e \in \mathcal E} \; \rho_e^{(t)}, \qquad t \in \{0, 1\},
$$
where $\mathcal E$ denotes the set of four causal experts. This refines the CATE prediction from Section~\ref{sec:experts} to use each group's selected expert:
$$
\hat\tau(\vx) = \hat y_1^{(e_1^\star)} - \hat y_0^{(e_0^\star)}.
$$

\begin{abox}
\textbf{Key Takeaway:} If a causal expert's embedding is valid, nearby units in that embedding should also have similar outcomes -- the idea underlying the variogram. The causal router scrutinizes this validity, independently per treatment group, and selects the expert for which it holds most strongly.
\end{abox}

 \section{Experiments} \label{sec:experiments}
\begin{table}[t]
\centering
\captionsetup{font=footnotesize}
\caption{\textbf{Performance comparison for CATE and ATE.} The results are reported as the mean and standard error across all realizations of a dataset. The \firstbest{best} and \secondbest{second best} methods are highlighted. The IPW method is not applicable to CATE.}
\label{tab:main}
\begin{adjustbox}{max width=\textwidth}
\LARGE
\begin{tabular}{lccccc|ccccc}
\toprule
\multirow{2}{*}{\textbf{Method}} &
\multicolumn{5}{c|}{\textbf{PEHE} $(\downarrow)$} &
\multicolumn{5}{c}{\textbf{ATE Relative Error} $(\downarrow)$} \\
\cmidrule(r){2-6}\cmidrule(l){7-11}
& IHDP & ACIC 2016 & \multicolumn{1}{c}{Lalonde {\footnotesize CPS}} & \multicolumn{1}{c}{Lalonde {\footnotesize PSID}} & \multicolumn{1}{c|}{Avg.} & 
IHDP & ACIC 2016 & Lalonde {\footnotesize CPS} & Lalonde {\footnotesize PSID} & \multicolumn{1}{c}{Avg.} \\
& & & ($\times10^{3}$) & ($\times10^{3}$) & Rank & & & & & Rank\\
\midrule

\rowcolor{bluesignal} 
\ours\,(ours)&
{0.80$\pm$0.15}&
{0.64$\pm$0.08}& 
\secondbest{9.05$\pm$0.02}& 
\firstbest{14.08$\pm$0.25}& 
\firstbest{2.50}&0.22$\pm$0.04& 
\secondbest{0.02$\pm$0.00}& 
{0.19$\pm$0.01}& 
\secondbest{0.12$\pm$0.02}& 
\firstbest{5.00}\\
\midrule

& \multicolumn{10}{c}{\textit{Tabular Foundation Models}} \\
TabPFN&
\firstbest{0.51$\pm$0.04}&
\firstbest{0.54$\pm$0.09}& 
{9.38$\pm$0.05}& 
{20.48$\pm$0.09}& 
{4.00}&0.20$\pm$0.04& 
\firstbest{0.01$\pm$0.01}& 
{0.24$\pm$0.17}& 
{0.71$\pm$0.18}& 
{6.50}\\

TabICL&
{0.65$\pm$0.09}&
{0.67$\pm$0.03}& 
{11.40$\pm$0.04}& 
{19.86$\pm$0.09}& 
{5.50}&0.20$\pm$0.04& 
{0.03$\pm$0.02}& 
{0.74$\pm$0.10}& 
{0.69$\pm$0.16}& 
{7.50}\\
\midrule

& \multicolumn{10}{c}{\textit{Causal Foundation Models}} \\
CausalPFN &
\secondbest{0.58$\pm$0.07}&
{0.92$\pm$0.11}&
\firstbest{8.96$\pm$0.02}&
\secondbest{14.40$\pm$0.20}&
\secondbest{3.50}& 0.20$\pm$0.04& 
{0.05$\pm$0.01}& 
\firstbest{0.13$\pm$0.01}& 
{0.22$\pm$0.02}& 
{6.25}\\

Do-PFN&
{6.06$\pm$0.94}&
{4.25$\pm$0.68}& 
{12.01$\pm$0.03}& 
{20.91$\pm$0.14}& 
{13.00}&0.57$\pm$0.95& 
{0.67$\pm$0.04}& 
{0.87$\pm$0.01}& 
{0.92$\pm$0.01}& 
{15.25}\\

CausalFM&
{8.71$\pm$0.90}&
{5.01$\pm$0.53}& 
{12.87$\pm$0.16}& 
{22.70$\pm$0.29}& 
{15.50}&1.24$\pm$0.01& 
{1.02$\pm$0.02}& 
{1.00$\pm$0.00}& 
{1.00$\pm$0.00}& 
{16.75}\\
\midrule

& \multicolumn{10}{c}{\textit{Traditional Causal Inference Models}} \\
T-Learner&
{1.73$\pm$0.30}&
{0.76$\pm$0.07}& 
{9.22$\pm$0.04}& 
{15.16$\pm$0.46}& 
{4.75}&0.21$\pm$0.04& 
{0.03$\pm$0.01}& 
{0.24$\pm$0.02}& 
{0.16$\pm$0.03}& 
\secondbest{5.75}\\

DA-Learner&
{2.07$\pm$0.36}&
{0.72$\pm$0.08}& 
{9.39$\pm$0.06}& 
{14.55$\pm$0.24}& 
{5.25}&0.23$\pm$0.04& 
{0.03$\pm$0.01}& 
{0.27$\pm$0.02}& 
{0.20$\pm$0.03}& 
{7.00}\\

DragonNet &
2.16$\pm$0.25& 
2.11$\pm$0.19& 
10.93$\pm$0.15& 
16.45$\pm$0.29& 
8.00& 0.20$\pm$0.04& 
0.06$\pm$0.02& 
0.55$\pm$0.03& 
0.47$\pm$0.03& 
7.75 \\

IPW &
—& 
—&
—&
—&
—& 0.24$\pm$0.04 & 
0.21$\pm$0.05 & 
\secondbest{0.17$\pm$0.01}& 
\firstbest{0.10$\pm$0.01}& 
{8.50}\\

RA-Net &
2.35$\pm$0.19 &
2.35$\pm$0.25 & 
11.74$\pm$0.09 & 
18.33$\pm$0.43 & 
9.50& 0.20$\pm$0.04 & 
0.07$\pm$0.03 & 
0.74$\pm$0.02 & 
0.50$\pm$0.04 & 
8.50 \\

X-Learner&
3.31$\pm$0.51& 
\secondbest{0.60$\pm$0.08}& 
12.15$\pm$0.15&
20.28$\pm$0.49& 
8.25&\secondbest{0.16$\pm$0.04}& 
0.03$\pm$0.01&
0.84$\pm$0.03&
0.72$\pm$0.03& 
6.75 \\

TarNet &
1.82$\pm$0.14& 
2.20$\pm$0.21& 
12.88$\pm$0.02& 
19.19$\pm$0.18& 
10.50& 0.20$\pm$0.04& 
0.05$\pm$0.02& 
1.00$\pm$0.00& 
0.78$\pm$0.01& 
10.50 \\

S-Learner &
2.57$\pm$0.41 &
0.85$\pm$0.13 & 
12.66$\pm$0.05 & 
21.80$\pm$0.18 & 
11.00&0.20$\pm$0.04 & 
0.03$\pm$0.01 & 
0.97$\pm$0.01 & 
0.90$\pm$0.02 & 
9.00 \\

BART &
2.50$\pm$0.39& 
{0.68$\pm$0.11}& 
12.81$\pm$0.05& 
21.36$\pm$0.16& 
10.00& 0.44$\pm$0.09& 
0.04$\pm$0.01& 
0.99$\pm$0.01& 
0.86$\pm$0.01& 
12.75 \\

GRF &
3.67$\pm$0.61& 
1.32$\pm$0.30& 
12.33$\pm$0.06& 
22.91$\pm$0.17& 
12.50& 0.18$\pm$0.03& 
0.07$\pm$0.02& 
0.82$\pm$0.02& 
0.85$\pm$0.02& 
10.00 \\

Forest DML&
4.53$\pm$0.73& 
1.48$\pm$0.31& 
12.95$\pm$0.04& 
22.99$\pm$0.15& 
15.00& \firstbest{0.08$\pm$0.01}&
0.05$\pm$0.01& 
1.03$\pm$0.01& 
1.05$\pm$0.01& 
11.50 \\

Forest DR Learner&
4.02$\pm$0.67&
1.34$\pm$0.29& 
15.98$\pm$0.68& 
22.78$\pm$0.54& 
14.25& 0.17$\pm$0.03&
0.04$\pm$0.02&
1.20$\pm$0.23& 
3.64$\pm$2.78& 
11.75 \\

\bottomrule
\end{tabular}
\end{adjustbox}
\end{table} \textbf{Experiment Setup.}
To evaluate \ours{}, we conduct experiments on four benchmark datasets -- IHDP~\citep{ramey1992infant}, ACIC 2016~\citep{dorie2019automated}, Lalonde CPS/PSID~\citep{lalonde1986evaluating} -- each generated with a different degree of semi-synthetic construction to produce known counterfactuals for evaluation (see Appendix~\ref{supp:datasets} for details).

We compare against three categories of methods. (1) \emph{Conventional causal estimators}: double machine learning (DML)~\citep{chernozhukov2018double, athey2019generalized, foster2023orthogonal}; the doubly robust learner (DR-Learner)~\citep{kunzel2019metalearners, kennedy2020optimal}; the T/S/X/DA learners~\citep{econml}; TarNet~\citep{shalit2017estimating}, DragonNet~\citep{shi2019adapting}, and RA-Net~\citep{curth2021inductive,curth2021catenets}; inverse propensity weighting (IPW)~\citep{rosenbaum1983central}; Bayesian additive regression trees (BART)~\citep{hill2011bayesian, chan2022data}; and generalized random forests (GRF)~\citep{athey2019generalized}. (2) \emph{Causal foundation models}: CausalPFN~\citep{balazadeh2026causalpfn}, Do-PFN~\citep{robertson2025pfn}, and CausalFM~\citep{ma2025foundation}. (3) \emph{General-purpose tabular foundation models}: TabPFN and TabICL. Performance is evaluated using the precision in estimation of heterogeneous effects (PEHE) for CATE and the relative ATE error for ATE (see Appendix~\ref{supp:metrics}).

\textbf{\ours{} Achieves Superior Performance on CATE and ATE.}
As shown in Table~\ref{tab:main}, the best and second-best methods vary considerably across datasets, and no single method wins consistently. The out-of-the-box tabular foundation models TabPFN and TabICL are themselves very competitive, achieving average ranks of 4.00 and 5.50 on CATE, and 6.50 and 7.50 on ATE, respectively. CausalPFN attains the second-best average rank (3.50) on CATE, but underperforms on ATE except on the Lalonde CPS dataset. 
\begin{wrapfigure}{r}{0.33\textwidth} 
    \vspace{-4mm}
    \centering
    \subfloat{\includegraphics[width=0.33\textwidth]{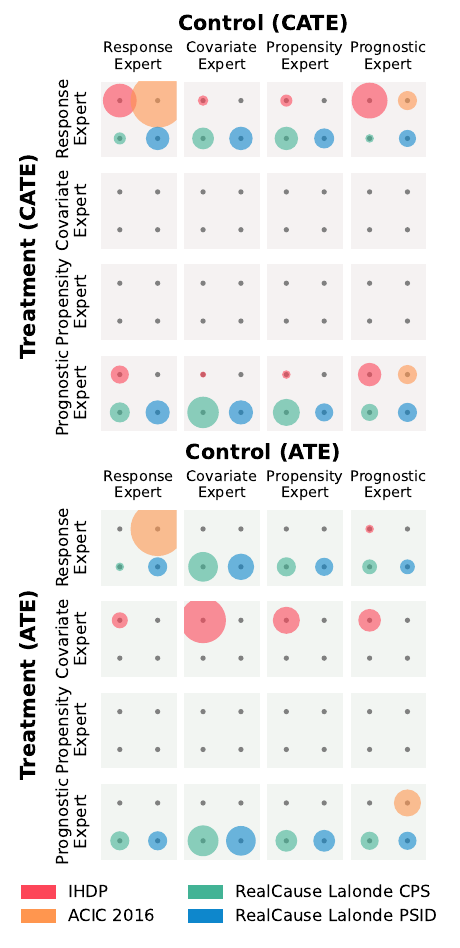}} 
\caption{Expert activation percentages for CATE (upper panel) and ATE (lower panel).}
	\label{fig:activation}
    \vspace{-6mm} 
\end{wrapfigure}
Traditional estimators can also achieve competitive, even best-in-class, performance in specific cases (e.g., IPW on the Lalonde datasets), but their performance is uneven across datasets, which lowers their average rank overall. \ours{} attains the best average ranks among all methods, 2.50 on CATE and 5.00 on ATE.

Average rank, however, only partially captures performance: when \ours{} is not the best or second-best (e.g., CATE on IHDP and ACIC), its performance remains on par with the top methods (Figures~\ref{fig:score_cate} and~\ref{fig:score_ate}). This absence of a "winner-take-all" method suggests that real-world SCMs are diverse enough that no single, fixed estimation strategy excels across all scenarios (Figure~\ref{fig:activation} shows this directly in the varying experts activated across datasets), underscoring the need for methods that adapt to the data at hand. Our approach opens a new avenue toward this end, without requiring a choice of evaluation metric on a held-out validation set. \ours{} also exhibits a better uncertainty quantification result (see Appendix~\ref{supp:uncertainty} for details).

\textbf{Analysis of Causal Experts and Router.}
For the MoE architecture to be meaningful, different experts should be selected across different datasets or tasks, and the selected expert should outperform both a randomly chosen expert and the simple average of all experts. We summarize several key findings toward this end.

(1) \emph{The variogram correlation provides an empirical signal for estimation performance.} As shown in Figure~\ref{fig:emb_cate1} (CATE on IHDP), for the Response Expert the absolute outcome difference increases almost monotonically with embedding distance, indicating that this expert captures the $Y \sim F(\mX, T)$ relationship along a lower-dimensional manifold; correspondingly, its PEHE is also the lowest among experts. In contrast, for experts with weaker or noisier distance-outcome relationships (e.g., the Prognostic Expert, whose variogram trend is positive but with higher variance), PEHE is correspondingly worse.

(2) \emph{For the same dataset, different tasks activate different experts.} Still on IHDP, but now for the ATE task, the Response Expert exhibits poor variogram correlation, and the Covariate Expert is selected instead (Figure~\ref{fig:emb_ate1}). This selection is corroborated by standalone evaluation: the Covariate Expert ($0.22 \pm 0.42$) outperforms the Response Expert ($0.28 \pm 0.48$) on ATE error, the reverse of their ranking on CATE (Table~\ref{tab:weighting}). We provide variograms for all the experts on all our datasets in Figures~\ref{fig:emb_cate1} to~\ref{fig:emb_ate4} for the inspection of different model behavior on different datasets.

(3) \emph{For the same dataset, different treatment arms activate different experts.} We plot expert activation percentages for control and treatment outcome estimation, across both CATE and ATE, in Figure~\ref{fig:activation}. Activation patterns vary substantially across datasets, arms, and tasks: for treatment-outcome estimation, only the Response and Prognostic Experts are ever activated, whereas for control-outcome estimation all four experts are activated, at differing rates. This indicates that \ours{} does not collapse onto a single expert across our benchmarks, but instead adapts its choice to the dataset, task, and arm.

(4) \emph{Expert activation may reflect (unknown) aspects of the data-generating process.} The RealCause Lalonde datasets are constructed semi-synthetically, with treatment and control groups artificially concatenated rather than drawn from a real experiment. Consistent with this, the Prognostic Expert (which ignores $T$) is activated at a higher rate on these datasets, suggesting a treatment-assignment process that differs from IHDP or ACIC (see Section~\ref{supp:datasets} for details). This suggests our causal embeddings, variogram, and routing mechanism may be useful for probing aspects of the underlying causal mechanism itself, particularly when prior knowledge about the DGP is limited.

\textbf{Ablation Experiments.}
(1) Our router is 1-sparse -- exactly one expert is selected per treatment arm for each dataset realization -- so for the router to be meaningful, it should outperform an unweighted average ensemble of all experts. Table~\ref{tab:ensemble} confirms this: the router substantially outperforms the ensemble, except on CATE for the Lalonde datasets, where the two are comparable. This corroborates that the routing mechanism removes the need to manually select a good, but a priori unknown, causal estimation strategy.
(2) In the causal router, we use Spearman's $\rho$ to measure the rank correlation between embedding distance and outcome difference. As a comparison, we evaluate alternative selection metrics -- Pearson correlation (linear) and the intrinsic dimension of the embedding itself -- both of which underperform Spearman's $\rho$ (Table~\ref{tab:router}). This indicates that the choice of routing metric matters for selecting the proper expert, and that Spearman's rank correlation is a more robust indicator than linear correlation or geometric dimensionality alone.
(3) \emph{Ablation on the expert architecture.} Each causal expert in \ours{} uses simplified attention with a learnable weighting MLP (Section~\ref{sec:topk}). Replacing this with a fixed $K$-Nearest-Neighbor (KNN) regressor generally performs worse (Table~\ref{tab:weighting}), confirming the value of the learned weighting function over uniform neighbor averaging.

\section{Discussion and Conclusion} \label{sec:discussion}
Causal effect estimation has traditionally required committing to a set of assumptions and an estimator largely independent of the dataset at hand, leaving the fit between assumption and data unexamined (Section~\ref{sec:related}). In this work, we integrate four causal estimation strategies -- Response, Covariate, Propensity, and Prognostic -- as parallel causal experts, so that no single method-specific assumption need be committed to a priori. A data-informed router, based on the distance-label variogram, then selects among these experts per dataset, task, and treatment arm, offering a more data-adaptive alternative to the fixed inference process of amortized causal foundation models.

In contrast to algorithm selection, we treat causal estimator selection as a dataset-dependent adaptation problem and show that characteristics observable directly from the data can inform which CATE estimation strategy is likely to perform well, without requiring a separate outcome-based validation set. Because our criterion is derived from the geometry of the data itself rather than from held-out predictive performance, it is also more directly interpretable -- one can inspect \emph{why} an expert was judged valid, rather than only observing that it scored well on a validation split. We discuss limitations and future direction in Appendix~\ref{sec:limitations}. The code is available at github.com/horsepurve/MoE-Learner.

% \section*{AI Use Statement}
% In this work, we used generative AI tools for grammar checking/correction and for retrieval and discovery (e.g., finding related work).
% We have not used generative AI tools for research ideation, execution, experimentation, or writing, and the rest of the required disclosure tasks are not applicable to this work.
% We have reviewed all AI-assisted work. 
% We take responsibility for the final content of this work, including text, claims or artifacts produced with the aid of generative AI.

% \subsection*{Reproducibility Statement}

% All the code, scripts, Jupyter Notebooks, and data files that are necessary for the reproduction of the work will be made publicly available on GitHub.

\bibliography{ref_main, ref_causalpfn}
\bibliographystyle{iclr2027_conference}

\appendix
\section{Implementation Details}

\subsection{Benchmarking Datasets} \label{supp:datasets}
All methods are tested on four benchmarking datasets, as shown in Table~\ref{tab:dataset_statistics}, with diverse numbers of covariates and sample sizes. All datasets are semi-synthetic but synthesis occurs at different stages of the construction. Their key characteristics and challenges are summarized as follows.  

\textbf{IHDP}~\citep{ramey1992infant} evaluates treatment effects of specialist home visits on the cognitive development of low birth-weight infants. 
It uses real pre-treatment covariates from a randomized experiment, while selection bias and potential outcomes are synthetically introduced using non-linear response surface equations. 
Its primary challenge stems from induced confounding coupled with a small sample size, making estimators prone to overfitting.

\textbf{ACIC 2016}~\citep{dorie2019automated}, derived from the Collaborative Perinatal Project, models health and demographic factors influencing maternal and child outcomes. 
The covariates are real, but the treatment assignments and potential outcomes are synthetically generated across 77 distinct data-generating processes. 
The main difficulty lies in handling extreme cross-setting diversity and high-dimensional noise, which severely tests whether causal overlap and unconfoundedness assumptions hold uniformly.

\textbf{RealCause Lalonde CPS}~\citep{lalonde1986evaluating} evaluates job training interventions by pairing experimental treated workers from the National Supported Work demonstration with a non-experimental comparison group from the Current Population Survey. 
A deep generative model is fitted on empirical data to synthesize joint distributions and ground-truth potential outcomes with known counterfactuals. 
Its main challenge is severe class imbalance and extreme propensity score polarization, where positivity assumptions break down for high-earning control subjects.

\textbf{RealCause Lalonde PSID}~\citep{lalonde1986evaluating} assesses the same job training program by combining the experimental treated cohort with observational controls from the Panel Study of Income Dynamics. 
Similar to the CPS benchmark, ground-truth outcomes are constructed via a fitted deep generative model to enable counterfactual evaluation. 
The primary hurdle is substantial distribution shift and severe positivity violations between treated and control groups, forcing models to rely on untestable structural extrapolation.
\begin{table}[th]
\centering
\begin{tabular}{c|c|c|c|c}
\toprule
\textbf{Dataset Group} & \textbf{IHDP} & \textbf{ACIC~2016} & \textbf{Lalonde {\tiny CPS}} & \textbf{Lalonde {\tiny PSID}} \\
\midrule
Realizations        & $100$ & $10$ & $100$ & $100$ \\
Data Sizes          & $747$ & $4802$ & $16177$ & $2675$ \\
Treated Group Sizes & \secondbest{$129$} & \secondbest{$1326$} & \secondbest{$156$} & \secondbest{$144$} \\
Control Group Sizes & $618$  & $3476$ & $16021$ & $2531$ \\
Treated Fractions   & $17.27\%$  & $27.61\%$ & $0.96\%$ & $5.38\%$ \\
\midrule
Covariate Dimensions  & $25$ & $58$ & $8$ & $8$ \\
\midrule
Training Data Sizes   & $672$ & $4321$ & $14559$ & $2407$ \\
Evaluation Data Sizes & $75$ & $481$ & $1618$ & $268$ \\
\bottomrule
\end{tabular}
\caption{Statistics for the benchmarking datasets used in the experiment. The numbers of the treated examples are \secondbest{highlighted}, showing a diverse range of percentages of the treated group.}
\label{tab:dataset_statistics}
\end{table}

\subsection{Evaluation Metrics} \label{supp:metrics}
Given a dataset of $N$ units with covariates and ground-truth CATE values
$\{(\vx^{(n)},\,\tau(\vx^{(n)}))\}_{n=1}^N$, and a
ground-truth ATE $\lambda$, we evaluate models using the relative ATE error and
the precision in estimation of heterogeneous effects (PEHE) \citep{hill2011bayesian}:
\begin{footnotesize}
\begin{equation}
\mathrm{Relative Error} = \frac{|\hat{\lambda}-\lambda|}{|\lambda|},
\qquad
\mathrm{PEHE} = \sqrt{\frac{1}{N}\sum_{n=1}^N
\bigl(\tau(\vx^{(n)})-\hat{\tau}(\vx^{(n)})\bigr)^2 }.
\end{equation}
\end{footnotesize}\\
Here, $\hat{\tau}$ and $\hat{\lambda}$ denote the estimated CATE and ATE, respectively.

\subsection{Causal Experts}

\begin{table}[t]
\centering
\captionsetup{font=footnotesize}
\caption{Comparison of \ours{} with averaging ensemble. Better results are  \secondbest{highlighted}.} 
\label{tab:ensemble}
\begin{adjustbox}{max width=\textwidth}
\LARGE
\begin{tabular}{lccccc|ccccc}
\toprule
\multirow{2}{*}{\textbf{Method}} &
\multicolumn{5}{c|}{\textbf{Mean PEHE} $(\downarrow \text{better})$} & \multicolumn{5}{c}{\textbf{Mean ATE Relative Error} $(\downarrow \text{better})$} \\ \cmidrule(r){2-6}\cmidrule(l){7-11}
& IHDP & ACIC 2016 & \multicolumn{1}{c}{Lalonde {\footnotesize CPS}} & \multicolumn{1}{c}{Lalonde {\footnotesize PSID}} & \multicolumn{1}{c|}{Avg.} & 
IHDP & ACIC 2016 & Lalonde {\footnotesize CPS} & Lalonde {\footnotesize PSID} & \multicolumn{1}{c}{Avg.} \\
& & & ($\times10^{3}$) & ($\times10^{3}$) & Rank & & & & & Rank\\
\midrule

\rowcolor{bluesignal} 
\ours\,(ours)&
\secondbest{0.80$\pm$0.15}&
\secondbest{0.64$\pm$0.08}& 
\secondbest{9.05$\pm$0.02}& 
\secondbest{14.08$\pm$0.25}& 
\secondbest{2.50}&\secondbest{0.22$\pm$0.04}& 
\secondbest{0.02$\pm$0.00}& 
\secondbest{0.19$\pm$0.01}& 
\secondbest{0.12$\pm$0.02}& 
\secondbest{5.00}\\
\midrule

Ensemble&
{2.35$\pm$0.86}&
{1.73$\pm$0.14}& 
\secondbest{9.04$\pm$0.21}& 
\secondbest{14.00$\pm$0.30}& 
{6.00}&0.26$\pm$0.46& 
{0.09$\pm$0.09}& 
{0.22$\pm$0.12}& 
{0.13$\pm$0.15}& 
{8.75}\\

\bottomrule
\end{tabular}
\end{adjustbox}
\end{table} \begin{table}[t]
\centering
\captionsetup{font=footnotesize}
\caption{Comparison of causal experts with ($\bigstar$) and without the MLP-based weighting function} 
\label{tab:weighting}
\begin{adjustbox}{max width=\textwidth}
\LARGE
\begin{tabular}{lcccccc|ccccc}
\toprule
\multirow{2}{*}{\textbf{Expert}} & \multirow{2}{*}{\makecell{Weighting\\Function}} &
\multicolumn{5}{c|}{\textbf{Mean PEHE} $(\downarrow \text{better})$} & \multicolumn{5}{c}{\textbf{Mean ATE Relative Error} $(\downarrow \text{better})$} \\ \cmidrule(r){3-7}\cmidrule(l){8-12}
&& IHDP & ACIC 2016 & \multicolumn{1}{c}{Lalonde {\footnotesize CPS}} & \multicolumn{1}{c}{Lalonde {\footnotesize PSID}} & \multicolumn{1}{c|}{Avg.} & 
IHDP & ACIC 2016 & Lalonde {\footnotesize CPS} & Lalonde {\footnotesize PSID} & \multicolumn{1}{c}{Avg.} \\
& & & & ($\times10^{3}$) & ($\times10^{3}$) & Rank & & & & & Rank\\
\midrule

\multirow{2}{*}{\makecell{Response\\Expert}}& $\bigstar$ &
\cellcolor{bluesignal}0.64$\pm$1.05 &\cellcolor{bluesignal}0.65$\pm$0.26 &\cellcolor{bluesignal}9.20$\pm$0.22 &\cellcolor{bluesignal}13.80$\pm$1.76 &\cellcolor{bluesignal}2.25 &\cellcolor{bluesignal}0.28$\pm$0.48 &\cellcolor{bluesignal}0.02$\pm$0.01 &\cellcolor{bluesignal}0.30$\pm$0.14 &\cellcolor{bluesignal}0.09$\pm$0.09 &\cellcolor{bluesignal}6.00 \\
&&
1.54$\pm$3.36 &0.97$\pm$0.37 &9.19$\pm$0.21 &13.75$\pm$1.76 &4.00 &0.27$\pm$0.48 &0.02$\pm$0.01 &0.30$\pm$0.14 &0.09$\pm$0.08 &6.00 \\
\midrule

\multirow{2}{*}{\makecell{Covariate\\Expert}}& $\bigstar$ &
\cellcolor{bluesignal}3.87$\pm$6.30 &\cellcolor{bluesignal}3.16$\pm$1.14 &\cellcolor{bluesignal}9.10$\pm$0.24 &\cellcolor{bluesignal}14.37$\pm$1.87 &\cellcolor{bluesignal}7.75 &\cellcolor{bluesignal}0.22$\pm$0.42 &\cellcolor{bluesignal}0.25$\pm$0.15 &\cellcolor{bluesignal}0.15$\pm$0.11 &\cellcolor{bluesignal}0.06$\pm$0.06 &\cellcolor{bluesignal}8.00 \\
&&
4.24$\pm$6.88 &2.82$\pm$1.43 &9.43$\pm$0.23 &14.66$\pm$1.72 &9.25 &0.23$\pm$0.43 &0.19$\pm$0.09 &0.16$\pm$0.11 &0.06$\pm$0.06 &7.75 \\
\midrule

\multirow{2}{*}{\makecell{Prognostic\\Expert}}& $\bigstar$ &
\cellcolor{bluesignal}0.85$\pm$1.54 &\cellcolor{bluesignal}1.31$\pm$0.67 &\cellcolor{bluesignal}9.23$\pm$0.21 &\cellcolor{bluesignal}13.92$\pm$1.91 &\cellcolor{bluesignal}4.25 &\cellcolor{bluesignal}0.27$\pm$0.50 &\cellcolor{bluesignal}0.08$\pm$0.06 &\cellcolor{bluesignal}0.32$\pm$0.13 &\cellcolor{bluesignal}0.11$\pm$0.12 &\cellcolor{bluesignal}9.50 \\
&&
1.62$\pm$3.26 &1.61$\pm$0.91 &9.23$\pm$0.21 &13.79$\pm$1.77 &5.00 &0.27$\pm$0.48 &0.08$\pm$0.05 &0.31$\pm$0.14 &0.10$\pm$0.11 &9.25 \\
\midrule

\multirow{2}{*}{\makecell{Propensity\\Expert}}& $\bigstar$ &
\cellcolor{bluesignal}5.25$\pm$8.05 &\cellcolor{bluesignal}3.09$\pm$1.65 &\cellcolor{bluesignal}9.04$\pm$0.25 &\cellcolor{bluesignal}14.96$\pm$4.83 &\cellcolor{bluesignal}8.75 &\cellcolor{bluesignal}0.26$\pm$0.46 &\cellcolor{bluesignal}0.19$\pm$0.23 &\cellcolor{bluesignal}0.16$\pm$0.14 &\cellcolor{bluesignal}0.38$\pm$0.39 &\cellcolor{bluesignal}9.25 \\
&&
5.28$\pm$8.12 &3.06$\pm$1.69 &9.01$\pm$0.21 &14.08$\pm$1.85 &8.25 &0.27$\pm$0.50 &0.11$\pm$0.05 &0.17$\pm$0.14 &0.33$\pm$0.15 &9.25 \\

\bottomrule
\end{tabular}
\end{adjustbox}
\end{table} \begin{table}[t]
\centering
\captionsetup{font=footnotesize}
\caption{Comparison of three metrics for the router. We estimate the intrinsic dimension using the maximum-likelihood estimator of Levina and Bickel~\citep{levina2004maximum}, which models the points in a local neighborhood as a homogeneous Poisson process and estimates the dimension from the distances to the $k$ nearest neighbors.} 
\label{tab:router}
\begin{adjustbox}{max width=\textwidth}
\LARGE
\begin{tabular}{lccccc|ccccc}
\toprule
\multirow{2}{*}{\textbf{Method}} &
\multicolumn{5}{c|}{\textbf{Mean PEHE} $(\downarrow \text{better})$} & \multicolumn{5}{c}{\textbf{Mean ATE Relative Error} $(\downarrow \text{better})$} \\ \cmidrule(r){2-6}\cmidrule(l){7-11}
& IHDP & ACIC 2016 & \multicolumn{1}{c}{Lalonde {\footnotesize CPS}} & \multicolumn{1}{c}{Lalonde {\footnotesize PSID}} & \multicolumn{1}{c|}{Avg.} & 
IHDP & ACIC 2016 & Lalonde {\footnotesize CPS} & Lalonde {\footnotesize PSID} & \multicolumn{1}{c}{Avg.} \\
& & & ($\times10^{3}$) & ($\times10^{3}$) & Rank & & & & & Rank\\
\midrule

\rowcolor{bluesignal} 
\textbf{Spearman's r}&
0.80±0.15 &0.64±0.08 &9.05±0.02 &14.08±0.25 &2.50 &0.22±0.04 &0.02±0.00 &0.19±0.01 &0.12±0.02 &5.00 \\

\textbf{Pearson's r}&
0.71±1.05 &0.63±0.27 &9.11±0.02 &14.56±0.87 &3.00 &0.22±0.42 &0.02±0.01 &0.25±0.16 &0.16±0.03 &5.50 \\

\textbf{Intrinsic Dimension}&
0.85±1.09 &1.27±0.82 &9.11±0.02 &14.37±0.87 &4.00 &0.27±0.48 &0.06±0.06 &0.15±0.01 &0.07±0.07 &7.50 \\

\bottomrule
\end{tabular}
\end{adjustbox}
\end{table} 
\textbf{Causal Embedding.} 
Pretrained on a vast collection of synthetic datasets and performing classification or regression at test time via in-context learning, TabPFN unifies heterogeneous datasets into a fixed-dimensional representation. 
For the causal embeddings, we use TabPFN version 8.2.0. Following the causal-expert construction described in Section~\ref{sec:experts}, TabPFN is used to embed each unit under the corresponding query-context relationship. To avoid label leakage when constructing these representations, we use 5-fold cross-validation: for each fold, the embedding of a unit is generated using a TabPFN model whose context does not contain that unit. Thus, the representation used for routing does not directly incorporate the unit's own observed outcome through the context. Empirically, we find that cross-validated embeddings perform better than embeddings generated without cross-validation.
TabPFN is only one possible choice for constructing the causal embeddings. Other tabular foundation models, such as TabICL, could be used within the same framework. We leave a systematic comparison and selection of the optimal embedding model for future work.

\textbf{Dissimilarity.} We instantiate $D_{ij}$ as cosine distance, $D(\vz_i, \vz_j) = 1 - \frac{\vz_i^\top \vz_j}{\|\vz_i\| \|\vz_j\|}$, which bounds $D \in [0, 2]$ and is invariant to feature norm, so retrieval depends only on directional alignment between query and context points.

\textbf{Monotonic weighting function.}
$f_\theta$ is a multilayer perceptron mapping the one-dimensional input $D$ (the cosine distance; equivalently, $1-D$ is the cosine similarity) to a scalar weight. Because larger similarity corresponds to smaller distance, the network first transforms the input to $1-D$. Every weight matrix $W^{(l)}$ is reparameterized as $\mathrm{Softplus}(\widetilde{W}^{(l)})$, ensuring all coefficients are non-negative, and Softplus activations are applied after every layer, including the output. Since a non-negative linear map composed with a non-decreasing activation is itself non-decreasing, the resulting network is non-increasing in $D$. The final Softplus guarantees $f_\theta(D) > 0$ everywhere, avoiding degenerate zero-weight neighborhoods. In our experiments $f_\theta$ uses one hidden layer of width $32$.

\textbf{Top-K retrieval.} For a context set of size $N$, exact top-$K$ selection is performed with a partial selection algorithm rather than a full sort, giving $O(N)$ average-case retrieval per query instead of $O(N \log N)$; this becomes relevant as $N$ grows beyond the scale used here.

\textbf{Training.} $f_\theta$ is trained on the training split itself: at each step we sample a random subset of $P$ points from $\mZ$ to serve as queries, and for each query retrieve its top-$K$ neighbors from the remaining $N - 1$ points under leave-one-out (the query point is masked out of its own context before retrieval, preventing the trivial shortcut $D(\vz, \vz) = 0$). The model is optimized with Adam to minimize mean squared error between $\hat{y}$ and the true label of each sampled query, $\mathcal{L} = \frac{1}{P}\sum_{p=1}^{P} (\hat{y}_p - y_p)^2$. Query subsets are resampled each training step. We use $K = 100$, $P = 400$, learning rate $0.1$, for $150$ steps.

\textbf{Inference.} At test time, the full training set $\mZ$ (all $N$ points) serves as the context; no leave-one-out masking is applied, since test queries are disjoint from $\mZ$.

\subsection{Causal Router}
\textbf{Representative-Point Sampling.}
To evaluate the geometry of each embedding efficiently, we compute the variogram statistics using a subset of maximally $M=150$ representative samples. We select these points using farthest-point sampling with cosine distance. Starting from an initial point, we iteratively select the point in $\mZ$ whose nearest selected point has the largest cosine distance, and continue until $M$ representative points have been obtained. This procedure encourages the representative set to cover diverse regions of the embedding space while avoiding an excessive concentration of samples in dense regions.

\textbf{Correlation-Based Routing.}
After selecting the representative points, we characterize the relationship between embedding distance and label variation using correlation measures. Specifically, we consider both Pearson's correlation and Spearman's rank correlation as empirical measures of the distance--label relationship. Pearson's correlation captures linear association, whereas Spearman's correlation captures monotonic association and is less sensitive to the precise functional form of the relationship.

In addition to these correlation-based measures, we compute the intrinsic dimension~\citep{levina2004maximum} of each embedding as a complementary geometric diagnostic. Intrinsic dimension characterizes the effective dimensionality of the manifold on which the embedding lies and has been shown to influence the convergence behavior of nearest-neighbor regression~\citep{kpotufe2011k,szekely2007measuring,anselin1995local}. A lower intrinsic dimension may therefore indicate that the embedding occupies a lower-dimensional structure, potentially making local retrieval more statistically efficient. However, intrinsic dimension alone does not determine local smoothness or predictive quality: an embedding can have low intrinsic dimension while still failing to organize samples according to their outcomes. We therefore treat intrinsic dimension as a complementary geometric property rather than a replacement for the distance--label correlation used by the router. As shown in Table~\ref{tab:router}, the Spearman's rank correlation is the most robust indicator for the expert performance.

\section{Proof of Theorem 1}
\textbf{Proof of Theorem~\ref{thm:variogram}} \label{app:proof}
Fix expert $e$ and, as in the main text, suppress the group subscript $t$ and expert superscript $(e)$; write $Z_i = \vz_i^{(e)}$ and $f = f_e$. Let $(Z_i, Y_i)$ and $(Z_j, Y_j)$ be independent draws from the same distribution, with $Y_i = f(Z_i) + \varepsilon_i$, $\E[\varepsilon_i \mid Z_i] = 0$, and $\operatorname{Var}(\varepsilon_i) = \sigma^2$.

Since $Y_i - Y_j = (f(Z_i) - f(Z_j)) + (\varepsilon_i - \varepsilon_j)$, squaring and conditioning on $D(Z_i, Z_j) = d$ gives

\begin{equation}
  \begin{split}
        V(d) &:= \E[(Y_i-Y_j)^2 \mid D_{ij}=d] \\
             &= \E[(f(Z_i)-f(Z_j))^2 \mid D_{ij}=d] \\
             &+ \E[(\varepsilon_i - \varepsilon_j)^2 \mid D_{ij}=d] \\
             &+ 2\,\E[(f(Z_i)-f(Z_j))(\varepsilon_i - \varepsilon_j) \mid D_{ij}=d].
  \end{split}
\end{equation}

The cross term vanishes since $\varepsilon_i, \varepsilon_j$ are conditionally mean-zero given $(Z_i, Z_j)$, and $\E[(\varepsilon_i-\varepsilon_j)^2] = \operatorname{Var}(\varepsilon_i) + \operatorname{Var}(\varepsilon_j) = 2\sigma^2$ by independence. 
Hence
\begin{equation}
  \begin{split}
    V(d) &= 2\sigma^2 + \E[(f(Z_i)-f(Z_j))^2 \mid D_{ij}=d], \\
    \gamma_e(d) &= \sigma^2 + \tfrac12\E[(f(Z_i)-f(Z_j))^2 \mid D_{ij}=d].
  \end{split}
\end{equation}
Since the second term is a conditional expectation of a nonnegative quantity, $\gamma_e(d) \ge \sigma^2$.

By $\beta$-local smoothness, almost surely
$$(f(Z_i)-f(Z_j))^2 \le L_e^2 D(Z_i,Z_j)^{2\beta},$$ 
so conditioning on $D_{ij}=d$ gives 
$$\E[(f(Z_i)-f(Z_j))^2 \mid D_{ij}=d] \le L_e^2 d^{2\beta}.$$ 

Substituting into the decomposition above yields 
$$\gamma_e(d) \le \sigma^2 + \tfrac12 L_e^2 d^{2\beta},$$ 
which together with the lower bound gives the claim. $\blacksquare$
 
\begin{figure}
    \centering
    \subfloat{\includegraphics[width=0.99\textwidth]{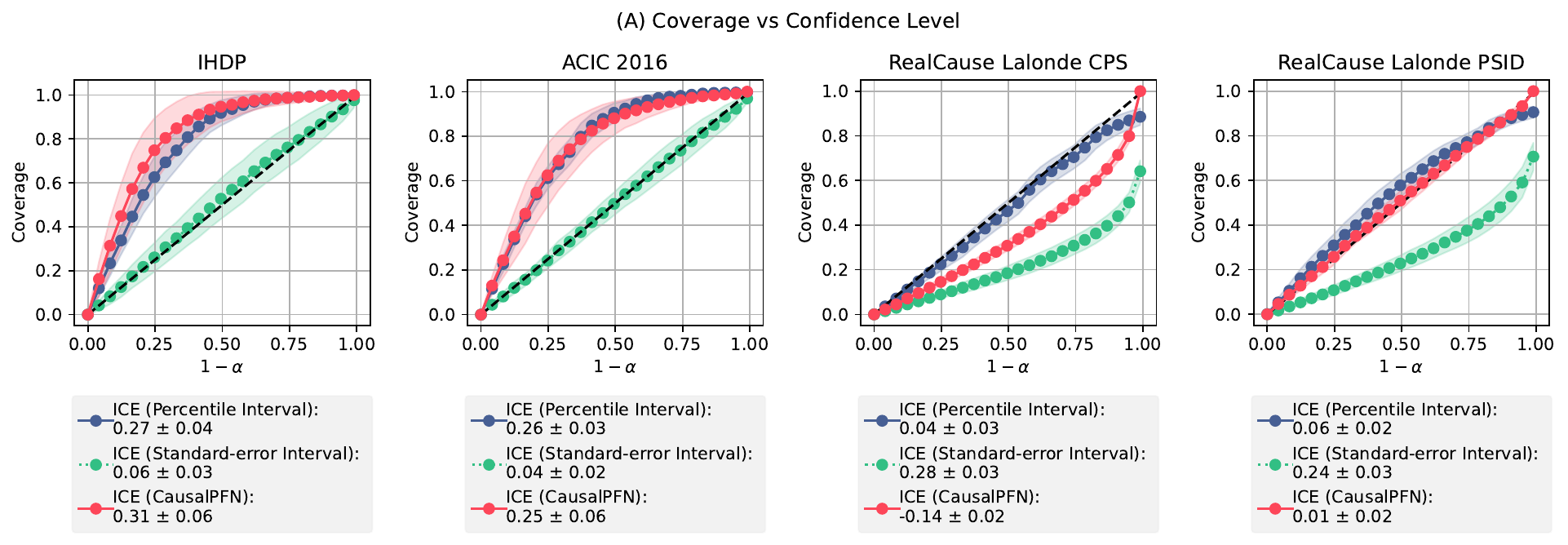}} \\
    \subfloat{\includegraphics[width=0.99\textwidth]{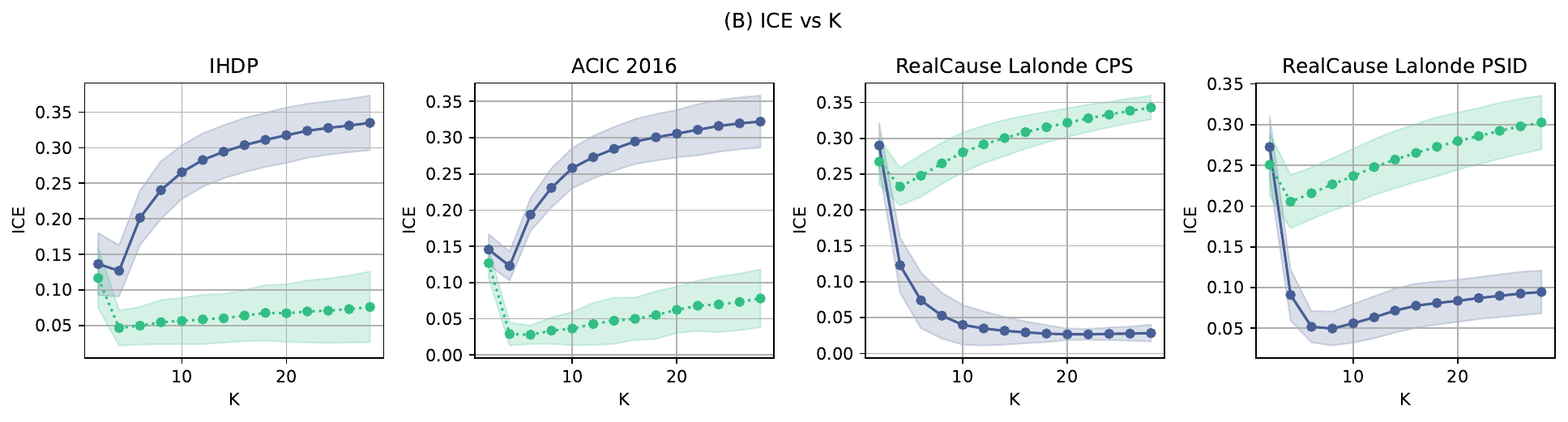}} \\
    \subfloat{\includegraphics[width=0.99\textwidth]{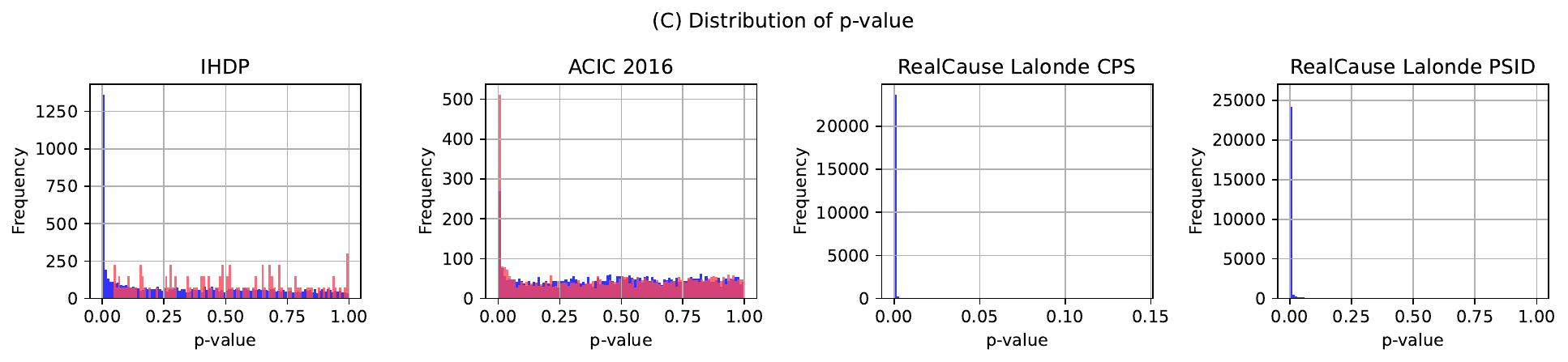}} 
    \caption{(A) ICE from the percentile interval, standard-error interval, and CausalPFN. (B) Sensitivity of ICE to $K$. (C) Different distribution of the p-values from Shapiro--Wilk test.}\label{fig:uncertainty}
\end{figure}
\section{Uncertainty Quantification} \label{supp:uncertainty}

Beyond a point estimate, the top-$K$ retrieval mechanism from Section~\ref{sec:topk} also supports uncertainty quantification, using the same neighborhood -- from each group's selected expert $e_t^\star$ -- already retrieved for prediction.

For each query point, recall the retrieved top-$K$ pairs $\{(\vz_t^{(k)}, y_t^{(k)})\}_{k=1}^{K}$ and their normalized attention weights $\{a_t^{(k)}\}_{k=1}^{K}$. We estimate the local variance from the retrieved labels as
$$
\hat{\sigma}_t^2 = \sum_{k=1}^{K} a_t^{(k)} \left(y_t^{(k)} - \hat{y}_t\right)^2,
$$
noting that the denominator $\sum_k a_t^{(k)}$ is omitted since the weights are already normalized to sum to one. We define the effective number of neighbors, following Kish's effective sample size, as
$$
\tilde{K}_t = \frac{1}{\sum_{k=1}^{K} \left(a_t^{(k)}\right)^2},
$$
which discounts $K$ toward the number of neighbors actually carrying weight when attention is concentrated on a few points. The standard error of the weighted prediction is then $\mathrm{se}(\hat{y}_t) = \hat{\sigma}_t / \sqrt{\tilde{K}_t}$.

Assuming the local label distribution is approximately Gaussian, a two-sided $(1-\alpha)$ uncertainty interval for $y_t$ is given by
$$
[L_t, U_t] = \left[\hat{y}_t - z_{1-\alpha/2}\,\mathrm{se}(\hat{y}_t),\; \hat{y}_t + z_{1-\alpha/2}\,\mathrm{se}(\hat{y}_t)\right],
$$
where $z_{1-\alpha/2}$ is the $(1-\alpha/2)$-quantile of the standard normal distribution.

When the local normality assumption is not supported, we instead construct an empirical percentile interval directly from the retrieved labels. For a nominal $(1-\alpha)$ coverage level, the interval is defined as
$$
[L_t, U_t] = \left[Q_{\alpha/2}\!\left(\{y_t^{(k)}\}_{k=1}^{K}\right),\; Q_{1-\alpha/2}\!\left(\{y_t^{(k)}\}_{k=1}^{K}\right)\right],
$$
where $Q_p(\cdot)$ denotes the empirical $p$-th percentile. This approach makes no Gaussian assumption, instead characterizing uncertainty directly from the observed local label distribution.

We construct a conservative interval for the CATE by subtracting the two group-level intervals:
\begin{equation}
[L, U] = [L_1 - U_0,\; U_1 - L_0]. \label{eq:cate-interval}
\end{equation}

We evaluate interval calibration using the Integrated Calibration Error (ICE), defined over a set of significance levels $\mathcal{A}$ as
$$
\mathrm{ICE} = \frac{1}{|\mathcal{A}|} \sum_{\alpha \in \mathcal{A}} \left|\widehat{\mathrm{coverage}}(\alpha) - (1-\alpha)\right|,
$$
where $\widehat{\mathrm{coverage}}(\alpha)$ is the empirical fraction of test examples for which the true $\tau_t$ falls within the corresponding interval $[L_t, U_t]$. Lower ICE indicates closer agreement between empirical and nominal coverage.

Unlike approaches such as TabPFN and CausalPFN, which obtain uncertainty estimates through repeated sampling or additional calibration procedures such as temperature scaling, our approach derives uncertainty directly from the local neighborhood already used for prediction, avoiding additional sampling or calibration overhead.
 
The uncertainty-quantification results are shown in Figure~\ref{fig:uncertainty}. We evaluate two types of confidence intervals: a standard-error-based interval and an empirical percentile interval. The choice between these intervals is motivated by the distribution of the retrieved outcomes within each local neighborhood.

To assess whether a normal approximation is appropriate, we perform a Shapiro--Wilk test on the retrieved outcome values for each local neighborhood. The null hypothesis is that the values within a neighborhood are normally distributed. A high $p$-value therefore provides no evidence against the normality assumption, whereas a $p$-value below $0.05$ leads us to reject it.

On IHDP and ACIC 2016, we observe consistently high $p$-values across local neighborhoods, indicating that the normality hypothesis is generally not rejected. In these settings, the standard-error-based intervals provide better ICE, achieving $0.06\pm0.03$ on IHDP and $0.04\pm0.02$ on ACIC 2016. In contrast, the Lalonde benchmarks exhibit markedly different behavior: for almost all realizations, the retrieved outcomes yield $p$-values below $0.05$, providing evidence against the normality assumption. Correspondingly, empirical percentile intervals achieve better ICE, with $0.04\pm0.03$ on Lalonde CPS and $0.06\pm0.02$ on Lalonde PSID.

These results support the use of a data-dependent choice of confidence-interval construction. In particular, when the retrieved outcomes are approximately compatible with a normal distribution, a standard-error-based interval is effective; when this approximation is not supported, empirical percentile intervals provide a more appropriate alternative.

\section{Limitations} \label{sec:limitations}
In this work we restrict our attention to method-specific assumptions under an unconfoundedness/back-door regime, and leave broader identification-strategy selection to future work. For example, settings where unconfoundedness itself fails call for different identification strategies altogether, such as Instrumental Variables (IV)~\citep{angrist1996identification}, Regression Discontinuity (RDD)~\citep{imbens2008regression}, or Difference-in-Differences (DiD)~\citep{bertrand2004how}.

As discussed in Section~\ref{sec:surrogate}, the variogram-based router is a surrogate diagnostic for local smoothness, not a direct certificate of causal assumption validity -- an embedding can appear smooth while still violating the assumption it is meant to encode, so routing decisions should be interpreted as data-informed evidence rather than formal guarantees. We also note that, unlike a standard MoE architecture for LLMs, our Mixture-of-Causal-Experts operates at the treatment-arm level, not the instance level, because the router must consider the local population of units. This points to future development of unit-level routing for more flexible causal inference.

We do not propose yet another causal foundation model; instead, we show how a general-purpose tabular foundation model can be retrofitted, via causal experts and a counterfactual-free causal router, into a causal effect estimator that outperforms both the underlying tabular foundation model and specialized causal foundation models. That said, Causal Foundation Models (e.g., CausalPFN, CausalFM, and Do-PFN) themselves are natively tabular transformer models, and can be used to extract various embeddings, which may possibly lead to better performance. We leave this question for future investigation.

\section{Additional Figures}
\begin{figure}
    \centering
    \subfloat{\includegraphics[width=\textwidth]{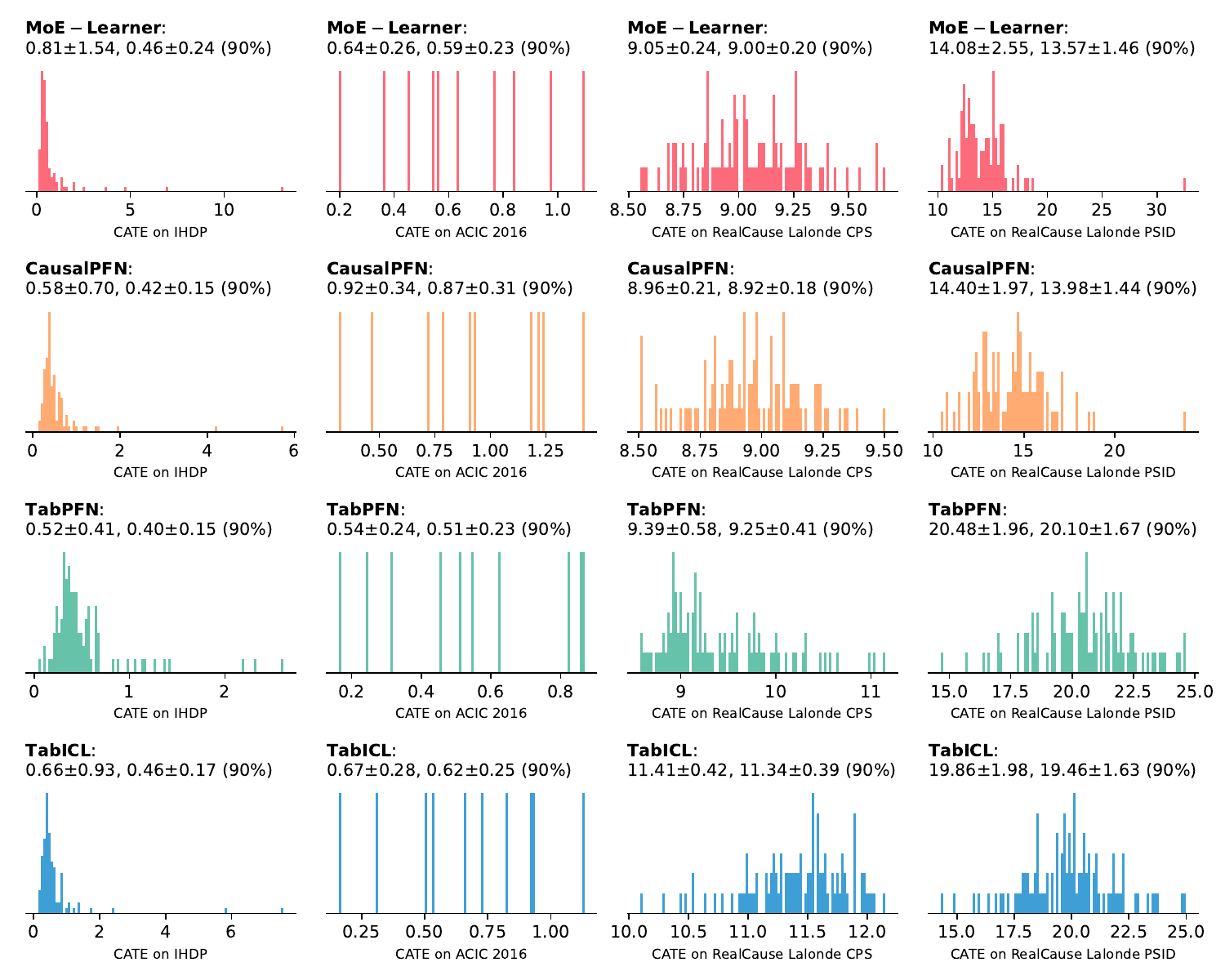}} 
    \caption{The distributions of PEHE scores from \ours{}, CausalPFN, TabPFN, and TabICL on four datasets, respectively. Lalonde PEHE is in thousands.}\label{fig:score_cate}
    \vspace{-1mm}
\end{figure}

\begin{figure}
    \centering
    \subfloat{\includegraphics[width=\textwidth]{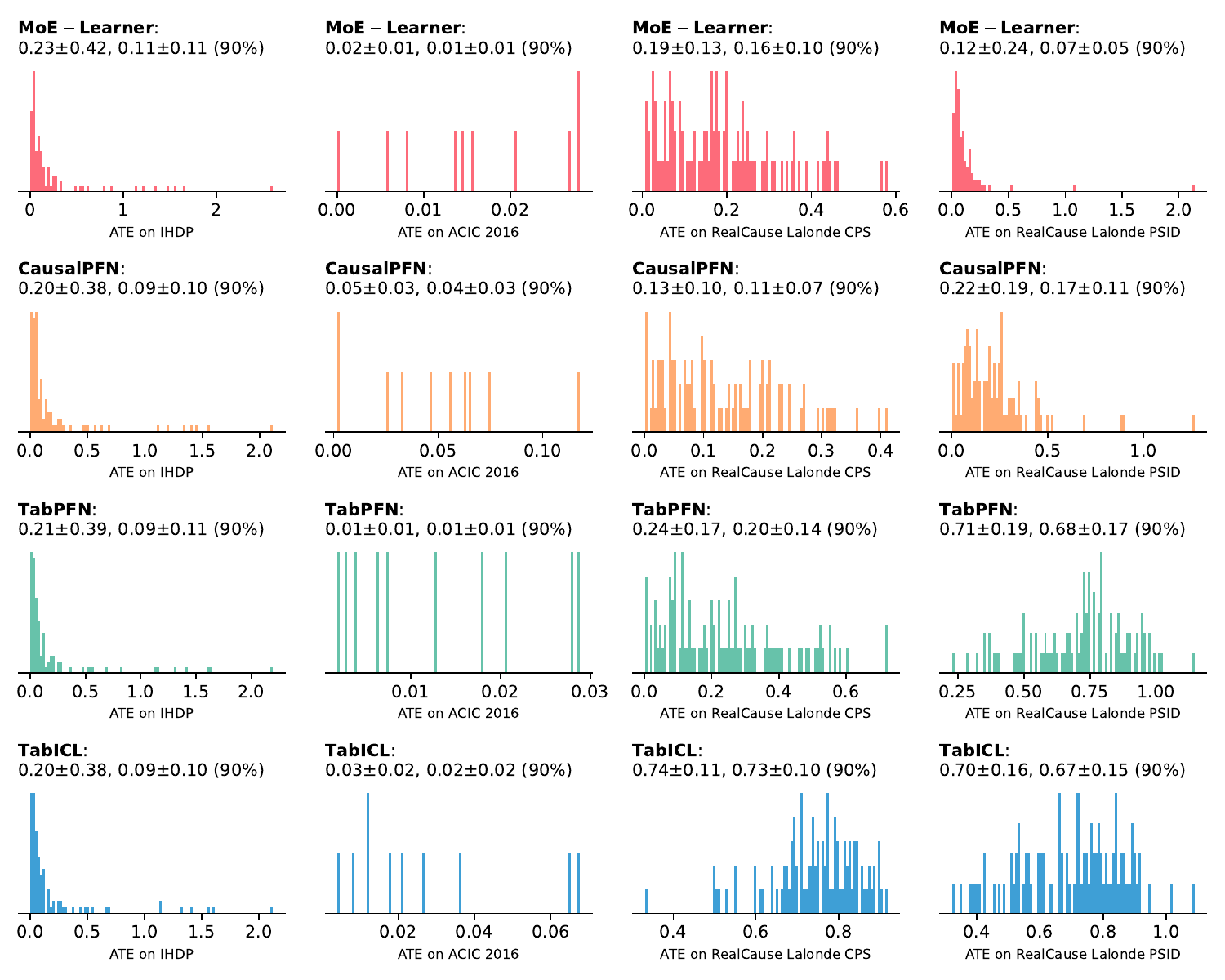}} 
    \caption{The distributions of the relative ATE errors from \ours{}, CausalPFN, TabPFN, and TabICL on four datasets, respectively.}\label{fig:score_ate}
    \vspace{-1mm}
\end{figure}

\begin{figure}
    \centering
    \subfloat{\includegraphics[width=\textwidth]{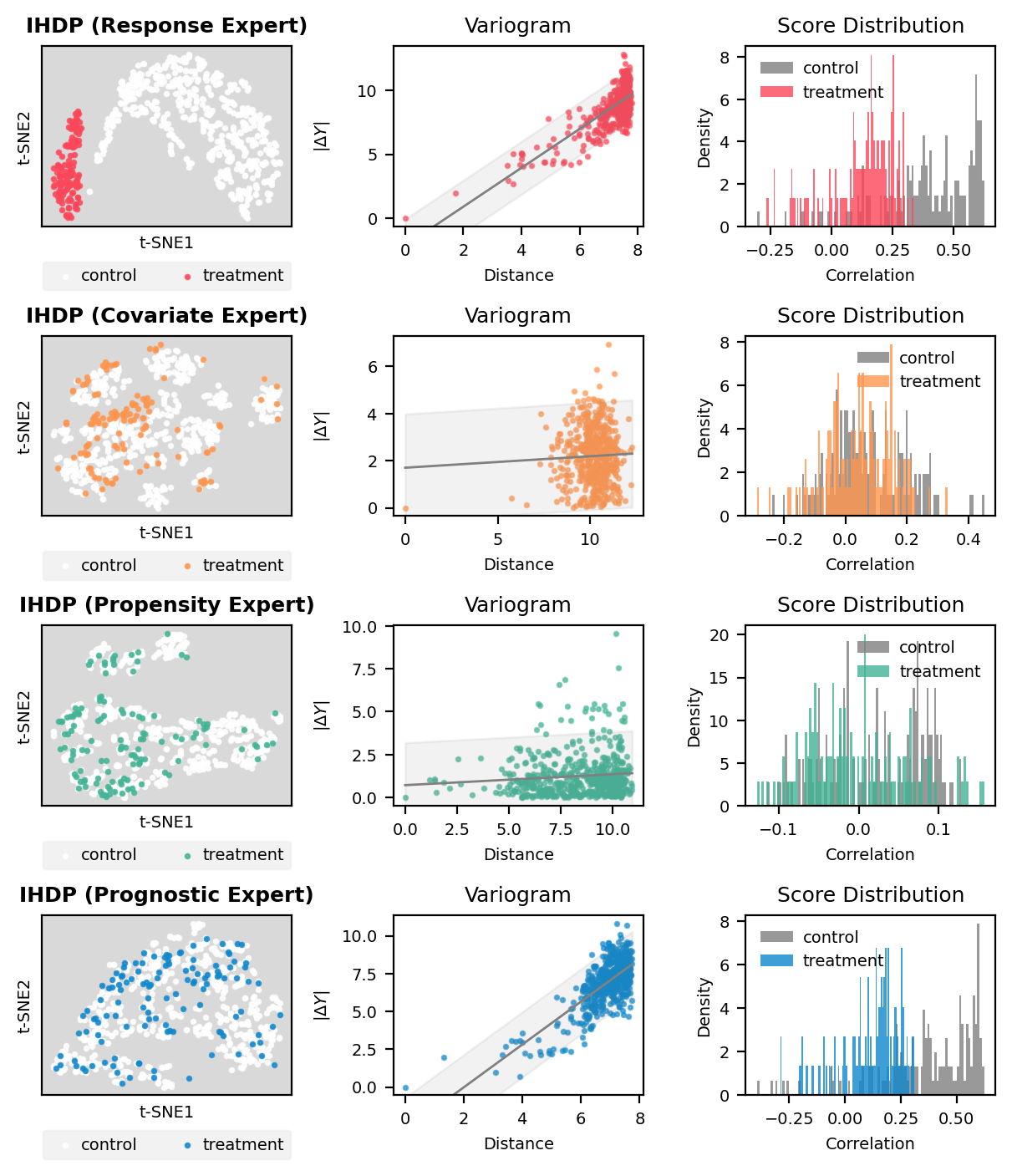}} 
    \caption{The t-SNE visualizations of embeddings (left column), the variograms (central column), and the distribution of the correlations in the variograms (right column) from the four experts on IHDP dataset for CATE.}\label{fig:emb_cate1}
    \vspace{-1mm}
\end{figure}

\begin{figure}
    \centering
    \subfloat{\includegraphics[width=\textwidth]{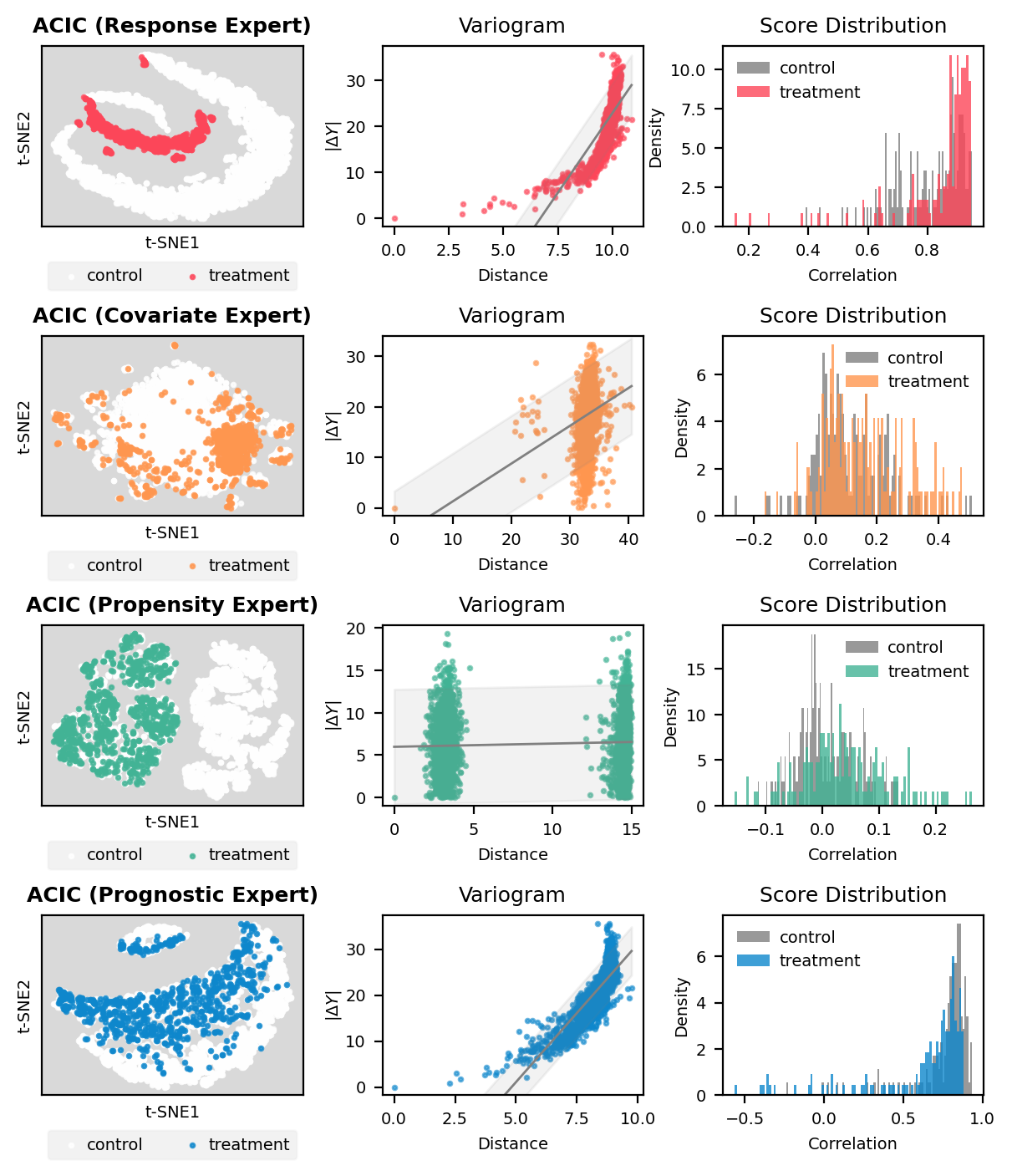}} 
    \caption{The t-SNE visualizations of embeddings (left column), the variograms (central column), and the distribution of the correlations in the variograms (right column) from the four experts on ACIC 2016 dataset for CATE.}\label{fig:emb_cate2}
    \vspace{-1mm}
\end{figure}

\begin{figure}
    \centering
    \subfloat{\includegraphics[width=\textwidth]{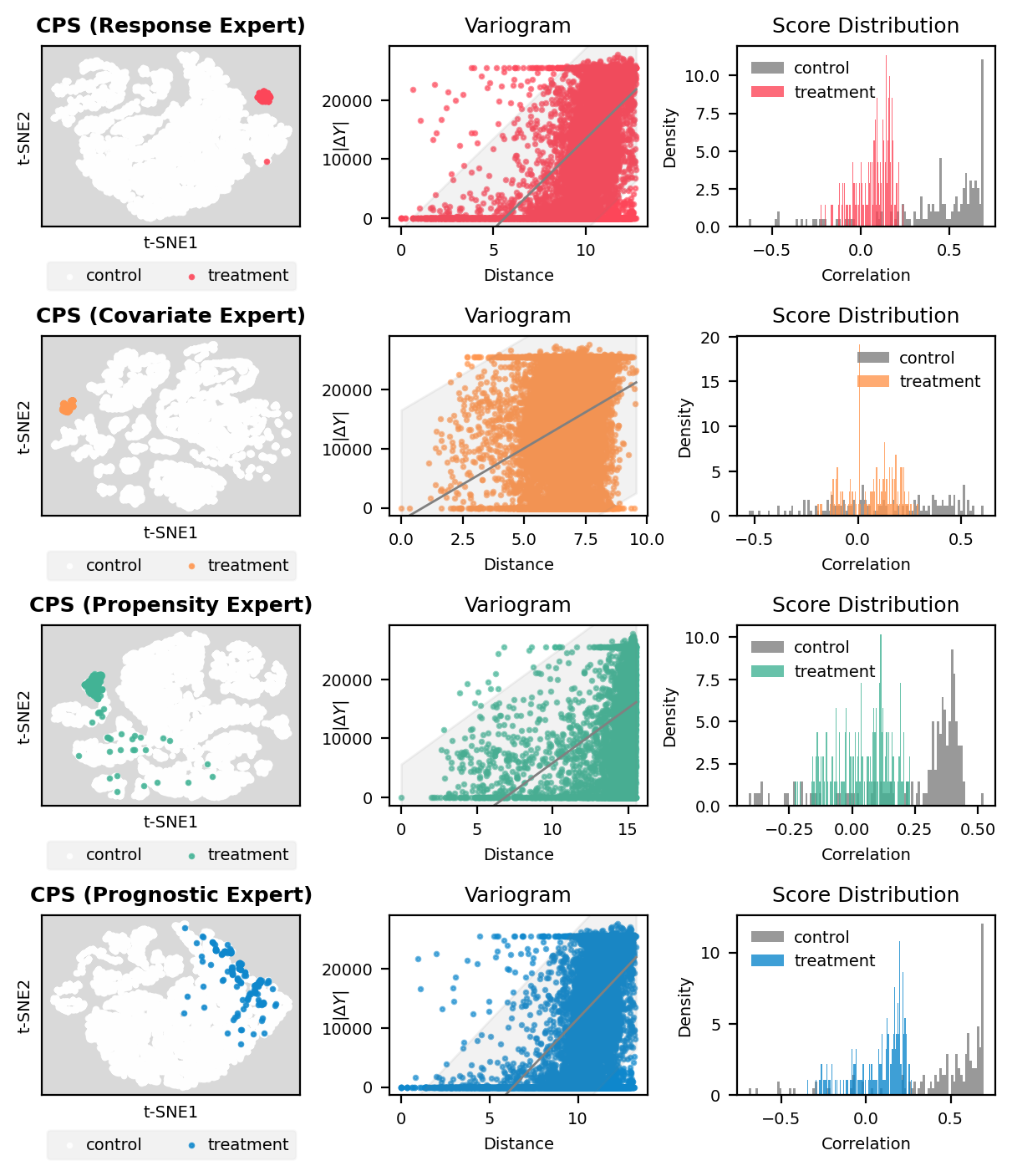}} 
    \caption{The t-SNE visualizations of embeddings (left column), the variograms (central column), and the distribution of the correlations in the variograms (right column) from the four experts on Lalonde CPS dataset for CATE.}\label{fig:emb_cate3}
    \vspace{-1mm}
\end{figure}

\begin{figure}
    \centering
    \subfloat{\includegraphics[width=\textwidth]{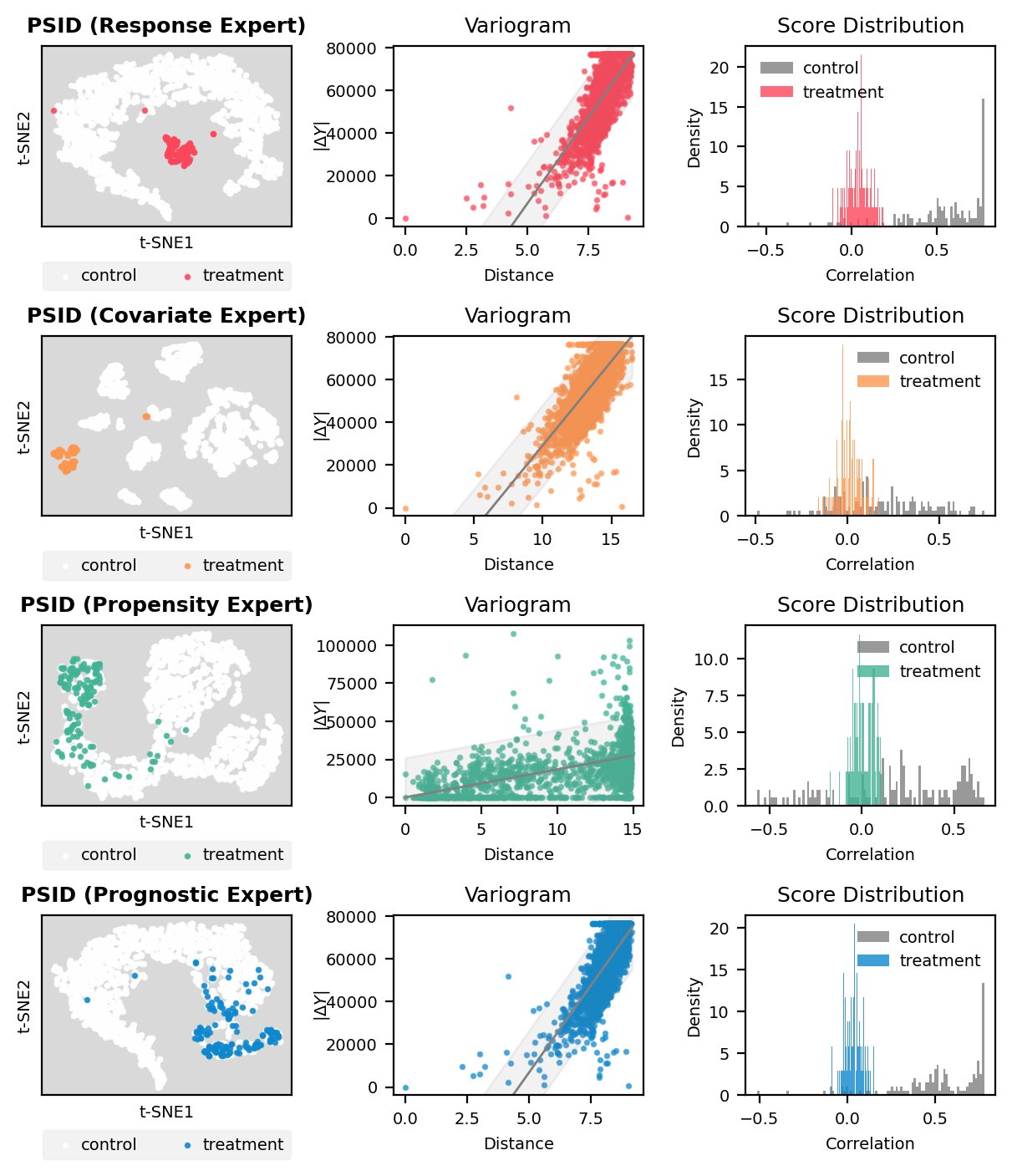}} 
    \caption{The t-SNE visualizations of embeddings (left column), the variograms (central column), and the distribution of the correlations in the variograms (right column) from the four experts on Lalonde PSID dataset for CATE.}\label{fig:emb_cate4}
    \vspace{-1mm}
\end{figure}

\begin{figure}
    \centering
    \subfloat{\includegraphics[width=\textwidth]{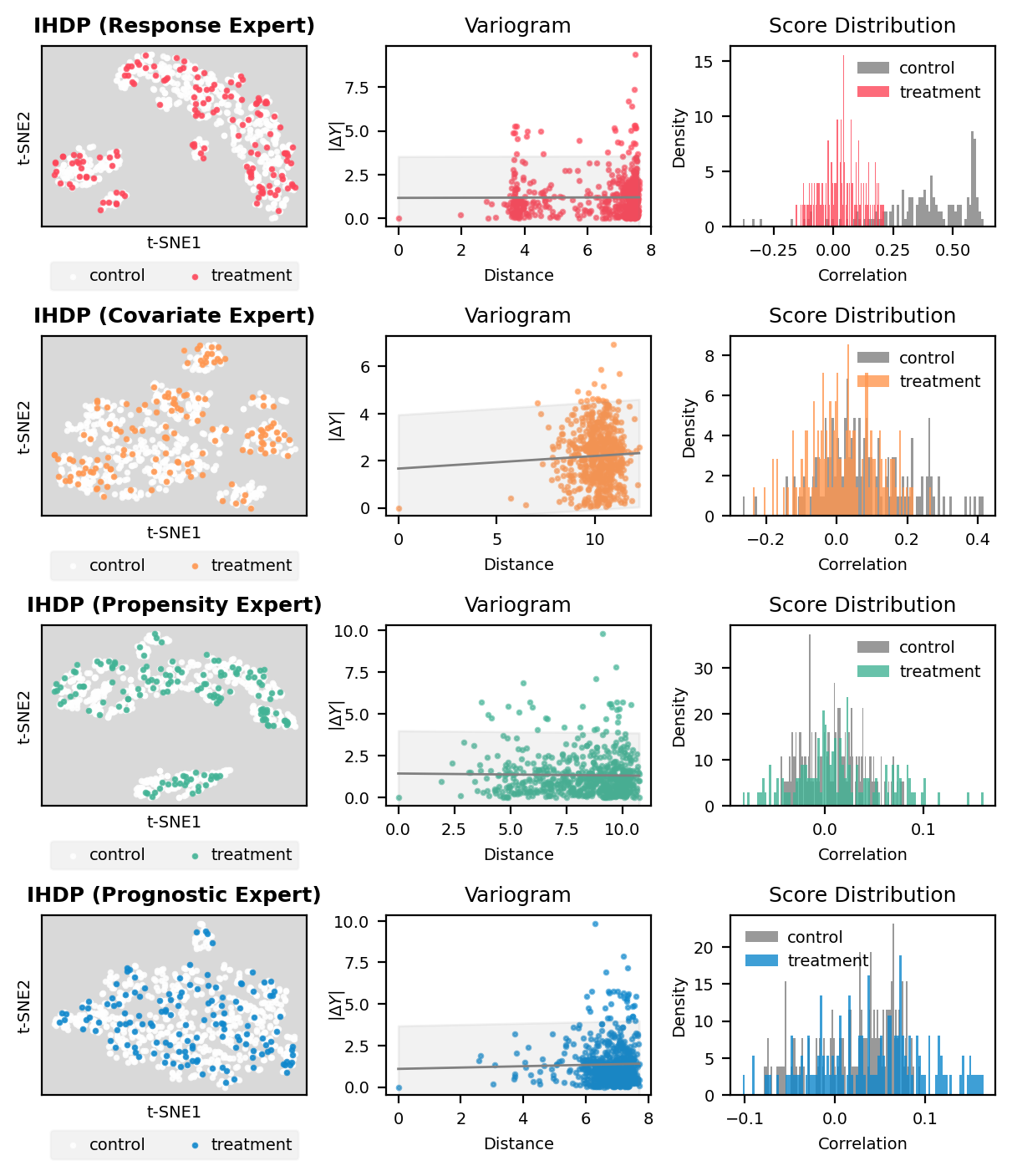}} 
    \caption{The t-SNE visualizations of embeddings (left column), the variograms (central column), and the distribution of the correlations in the variograms (right column) from the four experts on IHDP dataset for ATE.}\label{fig:emb_ate1}
    \vspace{-1mm}
\end{figure}

\begin{figure}
    \centering
    \subfloat{\includegraphics[width=\textwidth]{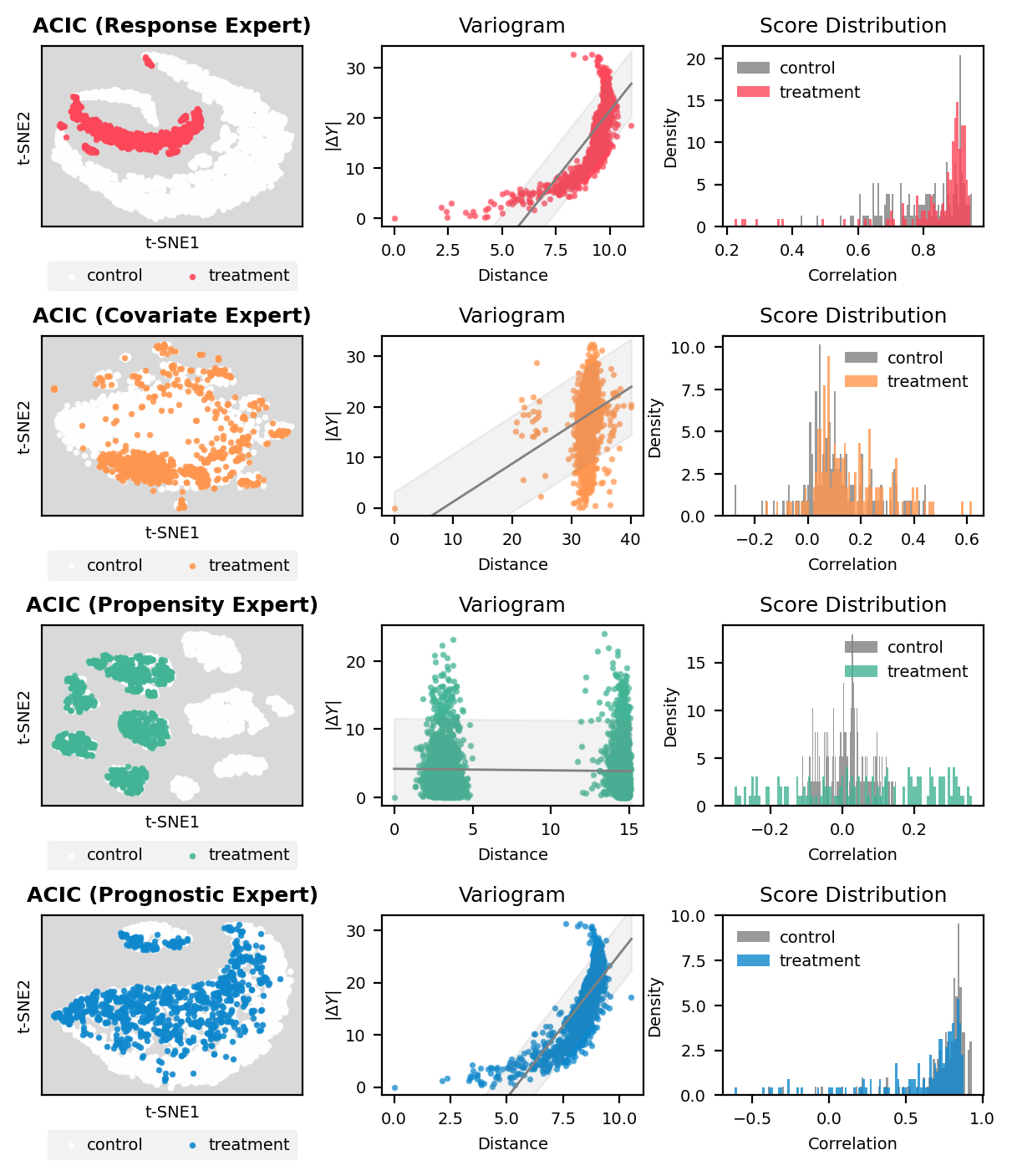}} 
    \caption{The t-SNE visualizations of embeddings (left column), the variograms (central column), and the distribution of the correlations in the variograms (right column) from the four experts on ACIC 2016 dataset for ATE.}\label{fig:emb_ate2}
    \vspace{-1mm}
\end{figure}

\begin{figure}
    \centering
    \subfloat{\includegraphics[width=\textwidth]{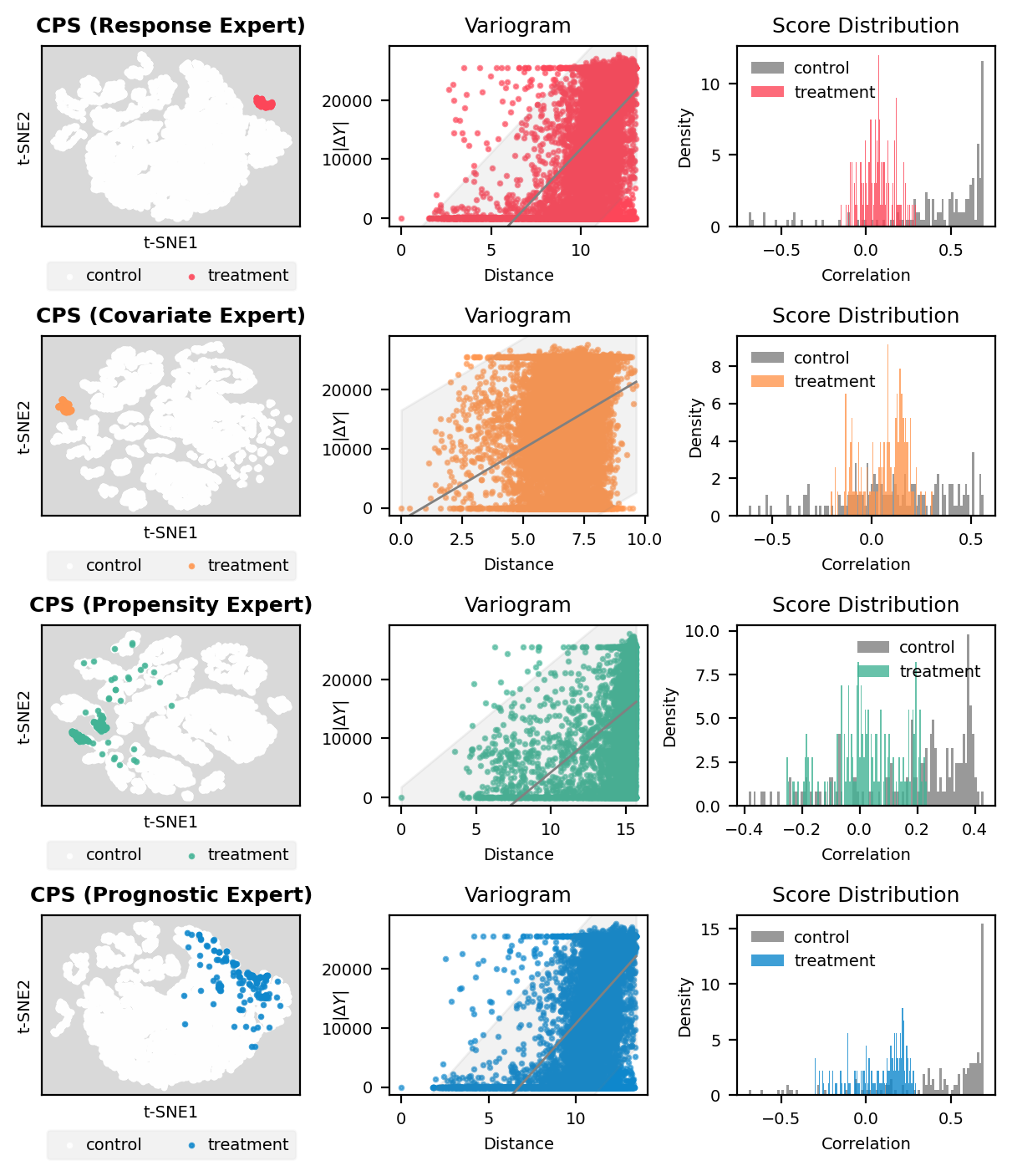}} 
    \caption{The t-SNE visualizations of embeddings (left column), the variograms (central column), and the distribution of the correlations in the variograms (right column) from the four experts on Lalonde CPS dataset for ATE.}\label{fig:emb_ate3}
    \vspace{-1mm}
\end{figure}

\begin{figure}
    \centering
    \subfloat{\includegraphics[width=\textwidth]{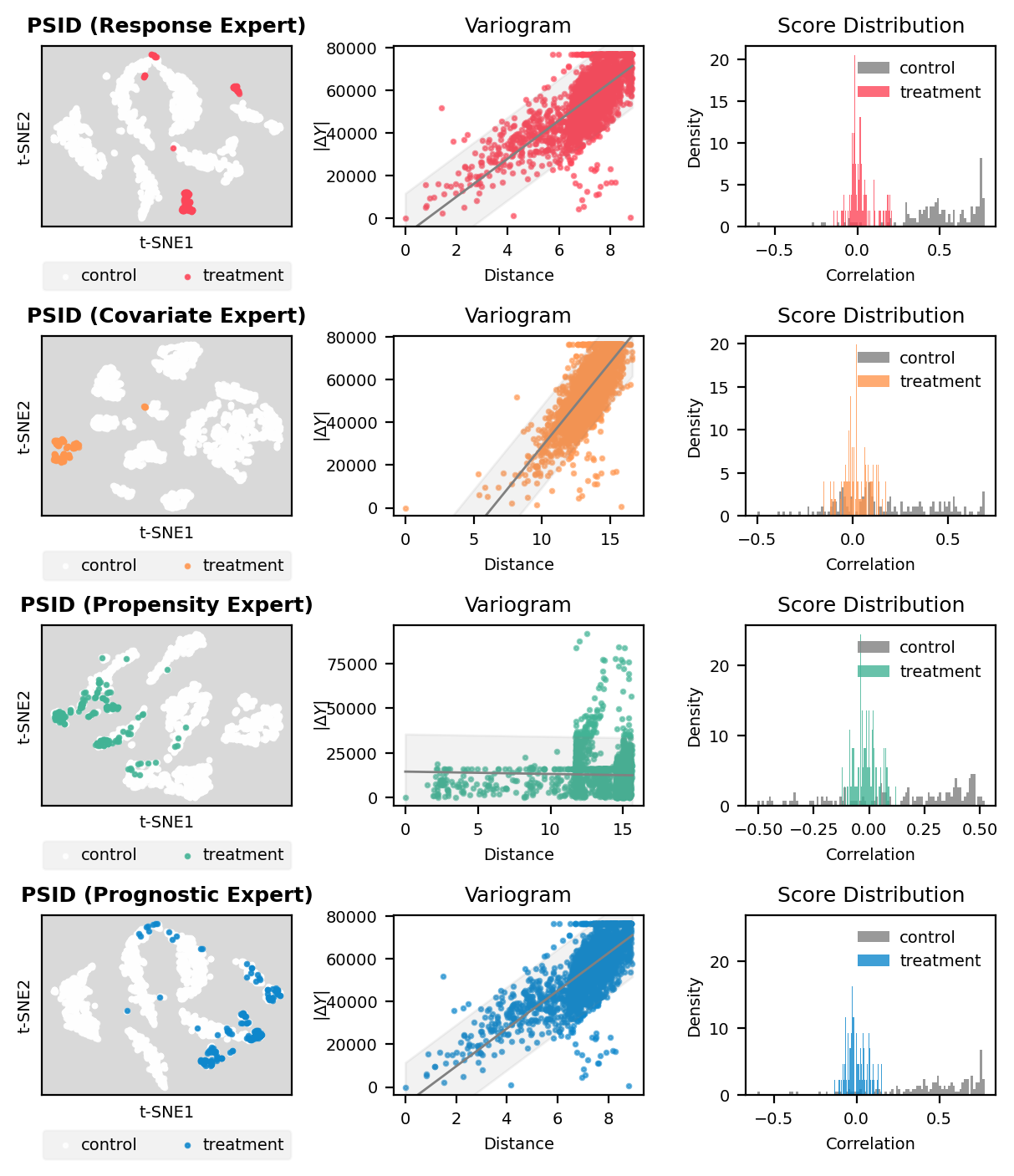}} 
    \caption{The t-SNE visualizations of embeddings (left column), the variograms (central column), and the distribution of the correlations in the variograms (right column) from the four experts on Lalonde PSID dataset for ATE.}\label{fig:emb_ate4}
    \vspace{-1mm}
\end{figure}  
\end{document}